\documentclass[manuscript,screen,nonacm]{acmart}

\AtBeginDocument{
  \providecommand\BibTeX{{
    \normalfont B\kern-0.5em{\scshape i\kern-0.25em b}\kern-0.8em\TeX}}}

\setcopyright{acmcopyright}
\copyrightyear{2020}
\acmYear{2020}
\acmDOI{XXX}
\acmBooktitle{Survey: Leverage Data Redundancy for DNN Optimizations}
\acmPrice{15.00}
\acmISBN{978-1-4503-XXXX-X/18/06}

\usepackage{tabularx} 
\usepackage[vlines]{tabularht}
\usepackage{adjustbox}
\usepackage{lscape}
\usepackage{graphicx}
\usepackage{multirow}
\usepackage{longtable}
\usepackage{makecell}
\usepackage{lipsum} 
\usepackage{afterpage}

\begin{document}

\title{Survey: Exploiting Data Redundancy for Optimization of Deep Learning}

\author{Jou-An Chen}
\email{jchen73@ncsu.edu}
\affiliation{
  \institution{Department of Computer Science, North Carolina State University}
  \country{USA}
}

\author{Wei Niu}
\email{wniu@email.wm.edu}
\author[3]{Bin Ren}
\email{bren@cs.wm.edu}
\affiliation{
  \institution{Department of Computer Science, William \& Mary}
  \country{USA}
}

\author{Yanzhi Wang}
\email{yanz.wang@northeastern.edu}
\affiliation{
  \institution{Department of Electrical and Computer Engineering, Northeastern University}
  \country{USA}
}

\author{Xipeng Shen}
\email{xshen5@ncsu.edu}
\affiliation{
  \institution{Department of Computer Science, North Carolina State University}
  \country{USA}
}

\authorsaddresses{This material is based upon work supported by the National Science Foundation (NSF) under Grants No. CCF-1525609, CCF-1703487, CNS-1717425, CNS-2028850, Department of Energy under Grant No. 20210107, and Jeffress Trust Awards in Interdisciplinary Research. Any opinions, findings, conclusions, or recommendations expressed in this material are those of the authors and do not necessarily reflect the views of NSF, DOE, or Thomas F. and Kate Miller Jeffress Memorial Trust. \\Authors’ addresses: Jou-An Chen, jchen73@ncsu.edu, Department of Computer Science, North Carolina State University, USA; Wei Niu, wniu@email.wm.edu; Bin Ren, bren@cs.wm.edu, Department of Computer Science, William \& Mary, USA; Yanzhi Wang, yanz.wang@northeastern.edu, Department of Electrical and Computer Engineering, Northeastern University, USA; Xipeng Shen, xshen5@ncsu.edu, Department of Computer Science, North Carolina State University, USA}

\renewcommand{\shortauthors}{Jou-An Chen et al.}
\newcommand{\TODO}[1]{{\it \color{red}\{TODO: #1\}}}
\newcommand{\rev}[1]{{\color{blue}#1}}

\begin{abstract}
Data redundancy is ubiquitous in the inputs and intermediate results of Deep Neural Networks (DNN). It offers many significant opportunities for improving DNN performance and efficiency and has been explored in a large body of work. These studies have scattered in many venues across several years. The targets they focus on range from images to videos and texts, and the techniques they use to detect and exploit data redundancy also vary in many aspects. There is not yet a systematic examination and summary of the many efforts, making it difficult for researchers to get a comprehensive view of the prior work, the state of the art, differences and shared principles, and the areas and directions yet to explore. This article tries to fill the void. It surveys hundreds of recent papers on the topic, introduces a novel taxonomy to put the various techniques into a single categorization framework, offers a comprehensive description of the main methods used for exploiting data redundancy in improving multiple kinds of DNNs on data, and points out a set of research opportunities for future to explore. 
\end{abstract}



\maketitle
\section{Introduction}
Deep learning, primarily powered by Deep Neural Networks (DNN), has repeatedly demonstrated its success and tremendous potential in revolutionizing the development of Artificial Intelligence and its applications in various domains. At a high level, DNN has two pillars: algorithm and data. DNN models embody the algorithm, and data is represented by DNN inputs and intermediate results (or called {\em activation maps}). Both algorithm and data are essential for the quality of a DNN, determining its accuracy, speed, size, energy efficiency, robustness, and so on. 

For DNNs, redundancy exists in both models and data. As many studies have shown, reducing some DNN layers or parameters or precision often leaves the model accuracy intact. The observation has prompted a large body of research in DNN compression, which tries to find effective ways to reduce the size of a DNN model and achieve more compact models or/and faster speeds. Many of the efforts resort to model pruning and quantization, while some compress models in the frequency domains~\cite{wang2016cnnpack} or optimize attentions in Transformer models~\cite{clark2019does, michel2019sixteen, voita2019analyzing}. 
We call those efforts \underline{model redundancy exploitations}. Several recent survey papers~\cite{cheng2017survey2, blalock2020state, ganesh2021compressing, choudhary2020comprehensive, hubara2017quantized, guo2018survey} have provided comprehensive overviews on the topic.

\underline{Data redundancy exploitation} is no less critical for DNN and has received much and increasing attention in recent several years as well. Many studies have noticed similar patches within an image, activation maps, or between adjacent video frames. Other studies have confirmed that some words in a sentence carry no significant meanings for the sentence. These observations have prompted many recent efforts in creating methods for detecting and leveraging the redundancy of various dimensions in all kinds of data. Hundreds of papers have been published, scattering in many venues over several years. But unlike model redundancy, there is not yet a systematic examination and summary of the many efforts, making it difficult for researchers to get a comprehensive view of the prior work, to learn about state of the art, to understand the differences and common principles among the many published studies, or to find out the areas and directions yet to explore. 

To the best of our knowledge, this paper offers the first one-stop resource for people interested in learning about studies on data redundancy for DNN. It provides a holistic view of the various techniques and their connections, offers insights on the limitations of state of the art, and potential directions and opportunities for the future to explore. The paper introduces the first known taxonomy on DNN data redundancy exploitation, putting the various topics' various techniques into a single categorization framework. It offers a comprehensive description of the main techniques used for detecting and exploiting data redundancy in improving multiple kinds of DNNs, from CNNs to RNNs, Transformers, and so on. It presents numerous data redundancy opportunities in images, videos, and texts and how existing studies tap into them with various techniques at a spectrum of granularities and scopes. It discusses the commonalities among the techniques, their differences, limitations, and promising research directions worth future explorations. 

We organize the rest of the paper as follows. Section~\ref{sec:scope} provides a formal definition of data redundancy and defines the scope of this survey. Section~\ref{sec:taxonomy} presents a taxonomy of studies on DNN data redundancy. Sections~\ref{sec:image},~\ref{sec:video}, and~\ref{sec:text} describe the practical techniques that prior work has developed in detecting and exploiting data redundancy for DNNs respectively on images, videos, and text data. Section~\ref{sec:future} discusses the limitations of existing explorations and points out several future directions. Section~\ref{sec:conclusion} concludes the survey with a summary.
\section{Terminology and Scope of Discussion} \label{sec:scope}

We start this section with clarifications of several terms essential for the rest of the paper, then define the scope of this survey, and explain the relations with several relevant concepts to our focus.

\subsection{Terminology}

\textbf{Activation Map.} In the traditional terminology, an \textit{activation map} (or called feature map) is the activations of filters applied to the input of a DNN also to the output of the previous layer. For the visual representation of an image or a single image frame in a video with RGB channels, it is a 3D tensor with $Height \times Width \times Channel$; for a video clip, it is a 4D tensor with $Height \times Width \times Channel \times Depth$, the \textit{Depth} is the number of frames. For ease of explanation, in this article, we extend the term \textit{activation map} to include the inputs to a DNN (the values of the input layer neurons).

\textbf{Hidden State.} For text representation, before input to the first layer of the model, the text is transformed into word vectors representation via a \textit{word embedding} (the collective name for a set of language modeling and feature learning techniques in natural language processing (NLP) where words or phrases from the vocabulary are mapped to vectors of real numbers~\cite{wordembedding}). Multiplied with weights, the output activations of a layer are called a \textit{hidden state} (which is also called \textit{encoder state}). It is represented as a 2D tensor with \textit{sequence length} $\times$ \textit{dimensionality of word embedding}. 

\textbf{Redundancy.} According to the dictionary~\cite{dict}, \textit{redundancy} is \textit{the state of being not or no longer needed or valuable}. It is hence a concept relative to a particular purpose. In the context of DNN, being useful usually refers to contributing to the quality of the DNN outputs, which is often measured by a certain kind of accuracy metric. \textit{Data redundancy} in DNN is hence defined as the data that is not or no longer useful for the quality of the outputs of DNN. {\em Data redundancy exploitation} for DNN refers to techniques that try to avoid data redundancy while keeping DNN outputs meeting the needs.

\subsection{Scope}
Data redundancy in DNN manifests in different kinds of forms. Regardless of the concrete structure, data redundancy in DNN must fall into one of the following three categories: (1)~\textbf{Repeated information}: Some part of the data conveys the identical (or similar) information as some other parts do. (2)~\textbf{Irrelevant information}: The information conveyed by some part of the data is irrelevant to the targeted outputs of the DNN. (3)~\textbf{Over-detailed information}: The information is conveyed in an unnecessarily detailed manner (e.g., image resolution). So any technique that tries to make a DNN avoid spending time and computations on any of the three types of information can be regarded as a technique for data redundancy exploitation. However, this general definition would blur the boundary between DNN algorithm, model, and data-centered optimizations. Some model optimizations (e.g., channel pruning~\cite{he2017channel, wang2018exploring, zhou2019accelerate,zhuang2018discrimination, hou2020feature, li2020feature, hou2020efficient,liu2018frequency}) are fundamentally driven by data redundancy (e.g., redundancy across channels). There are already surveys, particularly on model optimizations, but this survey focuses on data redundancy exploitation beyond model optimizations. These techniques help DNN avoid data redundancy in its computations to efficiently execute DNN in inference or training or eliminate noise to boost DNN accuracy. 

There is a body of work on considering data redundancy in the hardware design of DNN accelerators, such as cache buffer designs for data reuse in 2D CNNs~\cite{kim2020power, mocerino2019energy, jiao2018energy, ma2020axr, salamat2018rnsnet, hegde2018ucnn, wang2019none} and 3D CNNs~\cite{wang2020edge, wang2019systolic, fan2017f, wang2017enhanced, shen2018towards, hegde2018morph, chen20193d}, or integrating activation pruning into the hardware architecture designs~\cite{samal2020attention, piyasena2019reducing}. The techniques reduce the number of computations and bring speed and energy benefits. In this survey, we mainly focus our discussion on \emph{software-based techniques} for exploiting data redundancy.

\subsection{Relations with Relevant Concepts}
Several concepts are closely related to data redundancy exploitation. We next describe the relations with these concepts to further clarify our scope.

\textbf{Model Redundancy} Another aspect of redundancy in DNNs is model redundancy, referring to the redundancy within the parameters and architecture of a DNN. Model compression~\cite{cheng2017survey2, blalock2020state, ganesh2021compressing, choudhary2020comprehensive, hubara2017quantized, guo2018survey, adaptivecompression, acharya2018online, chen2016compressing}, including knowledge distillation~\cite{hinton2015distilling, urban2016deep}, parameter pruning, and weight quantization, are popular approaches to exploit model redundancy. This article focuses on \emph{data redundancy} of DNN, which refers to the redundancy within the input data of each DNN layer (usually in the form of multi-dimensional tensors).

\textbf{Data Preprocessing and Augmentation} Before data are fed into a DNN model, they often go through a preprocessing process. For training, that process is sometimes part of dataset augmentation~\cite{shorten2019survey, cubuk2019autoaugment}, increasing data variance to increase the models' generality. In computer vision, example operations include rotating, inverting, equalizing, Color changes, brightness adjustment, etc. The samples created by the transformations naturally carry some redundancy with the original data. For instance, with images inverted, two images contain the same set of pixel values, despite that the positions of individual image patches differ in the images. Such redundancy is implicitly considered when a technique exploits redundancy in input data. This survey exploits data redundancy and leaves data augmentation out of the scope.

\textbf{Feature Extraction and Data Embedding.} {\em Feature extraction} is a term in machine learning, referring to techniques that select or derive some features from a set of data that are regarded as most relevant to a specific machine learning task. Therefore, theoretically speaking, feature extraction can be viewed as a kind of exploitation of data redundancy as it reduces the raw data to an often smaller set of features. In a similar vein, {\em data embeddings} that map raw data to vectors in a space smaller than the original data space could also be regarded as a kind of reduction of data redundancy. Studies on these topics are usually considered as a separate research area named {\em feature extraction and representation}. They are typically not designed explicitly to exploit data redundancy, even though they may sometimes show such effects. We leave discussions on these studies out of the main focus of this paper.

\textbf{Dataset Selection.}
In the learning phase of DNN, \emph{training data selection}~\cite{fan2017learning, feng2019using, zheng2017training} is a process that tries to select the training data appropriate for the learner to achieve the learning task. To a certain degree, it may also remove some redundancy in the dataset (e.g., some data items in an over-sampled population). In this survey, we do not focus our discussion on this direction but on how data redundancy is addressed during the execution of the model (after the training dataset is selected or during inference). 
\section{Taxonomy} \label{sec:taxonomy}
Many studies detect or leverage data redundancy in DNN from several angles. A taxonomy that puts all the investigations into one categorization framework is essential for a holistic view of their differences, tradeoffs and shared principles. The taxonomy also offers a comprehensive view at the various dimensions of DNN data redundancy elimination, which can potentially guide the design of new redundancy elimination techniques.

\begin{figure}[h]
  \includegraphics[width=\textwidth,height=\textheight,keepaspectratio]{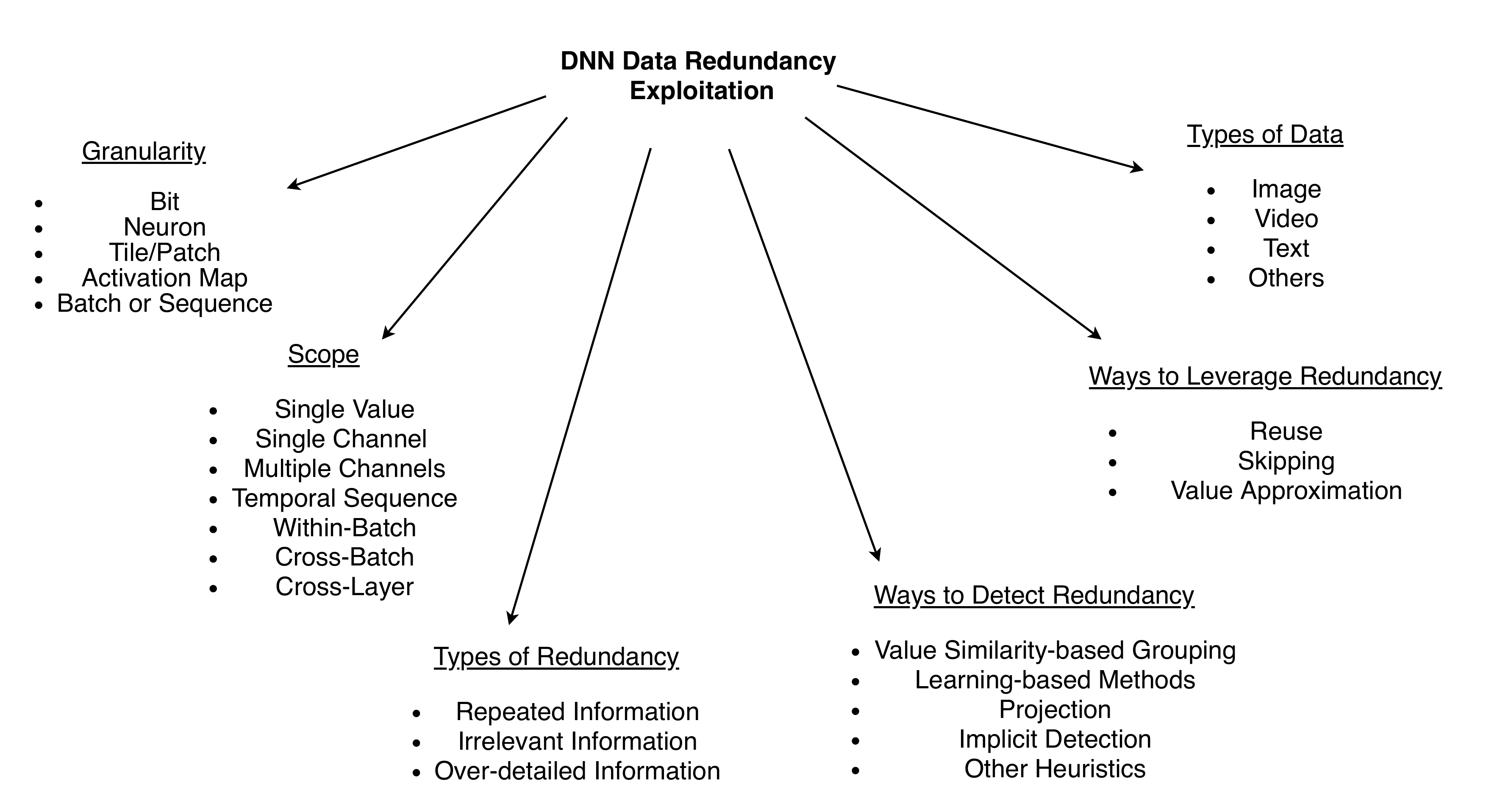}
  \caption{Taxonomy of data redundancy exploitation in DNN.}
  \label{fig:taxonomy}
\end{figure}

Figure~\ref{fig:taxonomy} presents a taxonomy we build after surveying hundreds of papers on DNN data redundancy exploitation. As far as we know, this is the first taxonomy on this topic. It shows the six most essential dimensions in DNN data redundancy exploitation. Data redundancy exploitation is usually the combination of one or more items in each dimension. We explain each of the dimensions next.

\subsection{Granularity}
This dimension refers to the unit of the data examined for data redundancy, which we will call {\em data unit} in the following discussions. There are five granularities listed below in increasing order in size. 
\begin{itemize}
    \item \textbf{Bit}: At this granularity, every bit in a value is the unit for examination. Data quantization and precision relaxation to the shorter representation of a value are commonly seen strategies exploiting bit-level redundancy. 
    Besides, bit representation can also be used directly for pattern-based bucketing. In RNSNet~\cite{salamat2018rnsnet}, for instance, an input value is transformed into its n-bits binary format and gets represented with a residue number system (RNS). The multiplications in neural networks are simplified to only addition and memory lookup, allowing memory-friendly operations.
    \item \textbf{Neuron}: The unit at this granularity is the value of a single neuron. An image is a pixel value at one channel or the value carried by a neuron in a DNN. 
    \item \textbf{Tile/Patch}: The values carried by a set of neurons, which correspond to a part of an activation map. It could be, for instance, a bounding box or region of interest (RoI) or channel in an image; in the case of text, it could be one or multiple sub-word embeddings. 
    \item \textbf{Activation Map}: The values in an entire activation map. 
    \item \textbf{Batch or Sequence}: A set of activation maps that the DNN carries at a layer, either at once or in order. The activation maps within the set can be either related or unrelated to each other. For example, sequential sentence input in a paragraph to a DNN can be correlated, while in a randomly shuffled set of training images, image activation maps are generally unrelated.
\end{itemize}

In general, the larger the granularity is, the less redundancy is there, but at the same time, the ratio between the benefits from removing a redundancy and the overhead in finding the redundancy is larger\footnote{Bit level is special, limited by the number of bits in one value, which is usually up to 32.}: Avoiding processing a whole batch of images saves more time than avoiding processing a neuron, but the chances in finding two identical batches are generally smaller than finding two identical neurons. On the other hand, the removals of different granularities of redundancy are not exclusive; one optimization could exploit redundancies at multiple granularities.


\subsection{Scope of Consideration}
This dimension is about the scope in which different data units are compared to identify similar units and hence redundancy. This dimension is related to the previous extent, granularity, but differ: For a given data unit, such as a patch of activation map, the comparisons for similar units can still vary (e.g., within one activation map, across activation maps in a batch, or batches). Specifically, the scopes considered in prior studies fall into one or more following. 

\begin{itemize}
    \item \textbf{Single Value}: Redundancy within a single representation of a value---the corresponding granularity is a bit.
    \item \textbf{Single Channel Activation Map}: In the case of an image, it refers to redundancy within a single channel activation map; in the case of text, it relates to redundancy within a single hidden state. The corresponding granularity can be either neuron or tile/patch. 
    \item \textbf{Multiple Channels of Activation Map}: This is for visual data (image and video) only. Redundancy across multiple channels of an activation map between neurons, patches/tiles, or single-channel activation maps.
    \item \textbf{Temporal Sequence of Activation Maps}: In the case of video, it refers to redundancy across a single video clip representation (4D tensor) between neurons, patches/tiles, single-channel activation maps, or activation maps. In the case of sequential text sequence input, it refers to redundancy across hidden states of the same sentence label (token type id), between neurons, sub-word embeddings, or single-sequence hidden states.
    \item \textbf{Within a Batch}: Inputs to a DNN are often provided in a batch each time. The scope for data redundancy consideration can be across inputs within a batch. 
    \item \textbf{Cross Batches}: The scope can also cross the boundaries of batches of inputs. 
    \item \textbf{Cross Layers}: All the earlier scopes typically assume that the comparisons are about activation maps at the same layer of a Neural Network. Some studies even expand the scope to activation maps on different layers of a Neural Network~\cite{park2019accelerating, dalvi2020exploiting}.
\end{itemize}

In general, a large scope subsumes a smaller scope: Redundancy discoverable in a smaller scope must be discoverable in a larger scope. But on the other hand, checking values in a larger scope also entails more overhead and complexity.  


\subsection{Types of Data Redundancy}
Data redundancy exploitation techniques can also be categorized based on what types of data redundancy they exploit. As earlier sections have mentioned, there are mainly three types of data redundancy. We list them here for the completeness of the discussion on the taxonomy.

\begin{itemize}
    \item \textbf{Repeated Information}: Similarity or duplication between data units. When the same operations operate on either identical or similar data units (e.g., multiplications on the same values, convolutions on similar image tiles), the results could be indistinguishable; the computations are redundant.
    \item \textbf{Irrelevant Information}: Unnecessary data units; if they are removed, will not harm the DNN computation results.
    \item \textbf{Over-detailed Information}: High-precision representation is not always necessary. For example, high-resolution images can be replaced with their low-resolution approximation in some application scenarios without causing the DNN to lose accuracy. Activation maps with lower precision sometimes give the same results for a DNN. A video with more frames than needed is another example. A sub-word vector with a higher-dimension representation (e.g., $1\times512$ versus $1\times256$ for a single word representation) or a higher precision value representation is also an example. 
\end{itemize}

The types of data redundancy define how redundancy appears. These different types of redundancy are complementary to each other; they require different ways to detect but at the same time can be capitalized together. Previous work has each focused on one of the three types of redundancy; it remains yet to investigate how to effectively combine the removals of multiple types of redundancy in one framework. Among the potential challenges, how to minimize the overhead is one of them.  

\subsection{Ways to Detect Data Redundancy}
The techniques can also be classified based on the ways they detect redundancy:

\begin{itemize}
    \item \textbf{Value Similarity-based Grouping}: This includes all kinds of similarity-based data grouping methods (e.g., various data clustering methods). 
    \item \textbf{Learning-based Methods}: These methods learn data redundancy on some data samples (often sampled from the training or validation datasets) and apply the knowledge at runtime execution of the model. An example is the learning-based quantization of activation maps.
    \item \textbf{Projection}: These techniques project the data representation to another domain space (e.g., frequency domain) and infer redundancy in that space.
    \item \textbf{Implicit Detection}: These techniques assume redundancy exists in a specific scope based on some prior knowledge about the domain. One of the examples is the techniques that leverage similarities between neighboring video frames.
    \item \textbf{Other Heuristics}: Methods that leverage some other domain-specific heuristics, such as the relative attentions or the average number of zeros appearing at a specific pixel location. 
\end{itemize}

The different ways apply to different domains and scenarios. The first two in the list are general, the third applies to the domains where projection across certain spaces makes sense, the fourth and fifth apply to domains where there is already certain prior knowledge about the redundancy in the domains.  

\subsection{Ways to Leverage Redundancy}
The different techniques take different ways to leverage data redundancy. They, however, in principle fall into the following three main categories.

\begin{itemize}
    \item \textbf{Reuse}: This approach reuses computation results on some data items for other data items. It applies to a similar kind of data redundancy. For example, Deep Reuse~\cite{ning2019deep} applies  \textit{Locality Sensitive Hashing (LSH)}~\cite{10.5555/645925.671516}, a fast unsupervised clustering scheme, to cluster data into similarity groups, and then uses the results computed on the cluster centroids as the results for all data items in the same cluster. 
    \item \textbf{Skipping}: This approach skips computations on some data items. It applies to data redundancy caused by irrelevant data items. 
    For instance, in Perforated CNN~\cite{figurnov2016perforatedcnns}, the authors deal with spatial redundancy by masking some pixels in CNN feature maps. It also includes techniques that select a subset of data representations. Some work does ad-hoc sampling, while some others make the selection in a more sophisticated way. In PoWER-BERT~\cite{goyal2020powerbert}, for instance, a retention configuration is defined to observe that cosine-similarity between vectors progressively increases as the layer goes deeper. Based on that setting, they employ strategies for word-vector selection. The retention configuration is further designed into learnable hyper-parameters.
    \item \textbf{Value Approximation}: This approach uses an approximation to generate data representation. It corresponds to data redundancy caused by over-detailed information. \textit{Quantization} and \textit{binarization} on data items fall into this category. Previous reviews~\cite{hubara2017quantized, simons2019review, qin2020binary} have provided comprehensive coverage on both research topics, so we do not emphasize them in the following discussion.
\end{itemize}

These three ways to leverage redundancy are complementary. Even though only the third in the list carries "approximation" in the name, all three could cause accuracy loss: {\em Reuse} may reuse results across similar but not identical data, and similarly, {\em skipping} may skip the computations of some similar but not identical data. These ways of leveraging redundancy can be used together. Some prior studies have already done that. One of them~\cite{dalvi2020exploiting}, for instance, exploits a \textit{layer selector} ("skipping") and \textit{correlation clustering} ("reuse") practice at the same time.

\subsection{Types of Data}
We can also classify the studies on data redundancy based on the types of data they deal with. Although there are many kinds of data, most existing studies on data redundancy are on the following three types of data:

\begin{itemize}
\item \textbf{Images.} This type of data includes various kinds of images, including those generated from higher dimension sensors (e.g., those collected through Lidar). 
\item \textbf{Videos.} This type of data features an extra-temporal dimension than images, which provides special opportunities for data redundancy exploitation.
\item \textbf{Texts.} This type of data consists of texts from various kinds of sources, usually written in Natural Languages.  
\end{itemize}

There are some other types of data that DNN has been applied to, such as graphs, genes, computer programs, and so on. We will briefly discuss them in Section~\ref{sec:future}. 

\subsection{Use of the Taxonomy} As far as we know, this is the first taxonomy on data redundancy exploitation. It was created based on our survey of hundreds of papers on data redundancy exploitation. The taxonomy can be used to position work in the big picture, to recognize the possible missing opportunities yet to explore for a certain kind of data, to guide the designs of future DNNs for efficiency, and to assist the possible creation of future automatic frameworks that may apply redundancy exploitation for new Deep Learning tasks. 

We categorize a set of representative papers listed in Table~\ref{tab:overview} using the taxonomy. We will discuss them in more detail in the next several sections. Specifically, we organize the following discussions based on several layers of the taxonomy. At the highest level, we divide the studies based on the types of data they handle into three sections, with Section~\ref{sec:image} on images, Section~\ref{sec:video} on videos, and Section~\ref{sec:text} on texts. We organize the various studies based on a dimension that captures the most prominent differences among the studies in each area. 

\begin{table}[h]
\caption{Some representative papers on data redundancy exploitation.}
\label{tab:overview}
\resizebox{1.\columnwidth}{!}{
\begin{tabular}{|l|l|l|l|l|l|l|}
\hline
\textbf{Data Type} & \textbf{Paper} & \textbf{Redundancy Type} & \textbf{Granularity} & \textbf{Scope} & \textbf{Detection} & \textbf{Exploitation} \\ \hline
\multirow{15}{*}{\textbf{Image}} &~\cite{ning2019deep} & Repeated & Patch/Tile & Multiple Channels, Within-Batch, Cross-Batch & Similarity & Reuse \\ \cline{2-7} 
 &~\cite{georgiadis2019accelerating} & Irrelevant, Repeated & Bit, Neuron & Single Value, Cross-Layer & Learning, Similarity & Value Approximation, Reuse \\ \cline{2-7} 
 &~\cite{park2019accelerating} & Repeated & Activation Map & Cross-Layer & Similarity & Reuse \\ \cline{2-7} 
 &~\cite{de2019skipping} & Repeated & Activation Map & Multiple Channels & Similarity & Reuse \\ \cline{2-7} 
 &~\cite{chen2019muffnet} & Repeated & Activation Map & Single Channel & Learning & Value Approximation, Reuse \\ \cline{2-7} 
 &~\cite{figurnov2016perforatedcnns} & Irrelevant & Neuron & Multiple Channels & Implicit Detection, Learning & Skipping \\ \cline{2-7} 
 &~\cite{akhlaghi2018snapea, hu2016network, shomron2019thanks} & Irrelevant & Neuron & Multiple Channels & Heuristic & Skipping \\ \cline{2-7} 
 &~\cite{chen2019layer, suzuki2018spectral} & Irrelevant & Neuron & Multiple Channels & Learning & Skipping \\ \cline{2-7} 
 &~\cite{gao2018channels} & Over-detailed & Activation Map & Multiple Channel & Learning & Skipping \\ \cline{2-7} 
 &~\cite{ibrokhimov2020effective} & Irrelevant & Neuron & Multiple Channels & Heuristic & Skipping (Subset Selection) \\ \cline{2-7} 
 &~\cite{lee2020subflow} & Irrelevant & Neuron & Cross-Layer & Heuristic & Skipping (Subset Selection) \\ \cline{2-7} 
 &~\cite{huang2018data} & Irrelevant & Neuron & Multiple Channels, Cross-Layer & Learning & Skipping (Subset Selection) \\ \cline{2-7} 
 &~\cite{chitsaz2020acceleration, chen2020frequency, chen2019drop} & Over-detailed & Activation Map & Multiple Channels & Projection & Value Approximation \\ \cline{2-7} 
 &~\cite{chen2020vip} & Over-detailed & Activation Map & Multiple Channels & Implicit Detection & Skipping (Ad-hoc Sampling) \\ \cline{2-7} 
 &~\cite{gao2018dynamic, li2019zoom} & Irrelevant & Patch/Tile & Multiple Channels & Learning & Skipping (Subset Selection) \\ \hline
\multirow{11}{*}{\textbf{Video}} &~\cite{kang2017noscope} & Irrelevant & Activation Map & Temporal Sequence & Similarity & Skipping \\ \cline{2-7} 
 &~\cite{cavigelli2017cbinfer, cavigelli2018cbinfer} & Irrelevant & Neuron & Temporal Sequence & Similarity & Skipping \\ \cline{2-7} 
 &~\cite{chin2019adascale} & Over-detailed & Patch/Tile & Multiple Channels & Learning & Value Approximation \\ \cline{2-7} 
 &~\cite{zhang2017kill} & Irrelevant & Patch/Tile & Multiple Channels & Implicit Detection & Skipping (Subset Selection) \\ \cline{2-7} 
 &~\cite{mao2018catdet} & Repeated & Patch/Tile & Temporal Sequence & Implicit Detection & Skipping (Subset Selection) \\ \cline{2-7} 
 &~\cite{yeung2015endtoend, alwassel2017action, wu2018adaframe, wu2019multiagent, korbar2019scsampler} & Over-detailed & Activation Map & Temporal Sequence & Learning & Skipping (Subset Selection) \\ \cline{2-7} 
 &~\cite{su2016leaving} & Over-detailed & Batch or Sequence & Temporal Sequence & Learning & Skipping (Subset Selection) \\ \cline{2-7} 
 &~\cite{zhu2016deep, zhu2017flowguided, zhu2017high} & Repeated & Activation Map & Temporal Sequence & Implicit Detection & Reuse (Temporal Propagation) \\ \cline{2-7} 
 &~\cite{Kang_2017, Kang_2018} & Repeated & Activation Map & Temporal Sequence & Implicit Detection & Reuse (Temporal Propagation) \\ \cline{2-7} 
 &~\cite{chen2018optimizing} & Repeated & Activation Map & Temporal Sequence & Implicit Detection & Reuse (Temporal Propagation) \\ \cline{2-7} 
 &~\cite{liu2017mobile} & Repeated & Activation Map & Temporal Sequence & Implicit Detection & Reuse (Temporal Propagation) \\ \hline
\multirow{5}{*}{\textbf{Text}} &~\cite{liu2018generating, dai2020funneltransformer} & Over-detailed & Patch/Tile & Activation Map & Implicit Detection & Skipping (Ad-hoc Sampling) \\ \cline{2-7} 
 &~\cite{goyal2020powerbert} & Irrelevant & Patch/Tile & Activation Map & Learning & Skipping (Subset Selection) \\ \cline{2-7} 
 &~\cite{kitaev2019reformer} & Repeated & Patch/Tile & Activation Map & Similarity & Reuse \\ \cline{2-7} 
 &~\cite{dalvi2020exploiting} & Repeated, Irrelevant & Batch or Sequence & Temporal Sequence, Cross-Layer & Similarity & Reuse, Skipping (Subset Selection) \\ \hline
\end{tabular}
}
\end{table}

In Section~\ref{sec:image}, we divide studies on image data redundancy based on the types of redundancy they are dealing with, as the number of works in the three sub-categories (\textit{Repeated Information}, \textit{Irrelevant Information}, and \textit{Over-detailed Information}) are relatively balanced and offers comprehensive coverage of the differences among the techniques. We point out the discrepancies between the work in each group in other dimensions (e.g., data representations, redundancy exploitation techniques) when necessary during the discussion. 
While many image redundancy exploitation techniques could potentially be applied to each frame in a video, Section~\ref{sec:video} focuses on the methods specially designed for videos. These efforts extend the work in Section~\ref{sec:image} by exploring how redundancy is leveraged on the patch unit and the temporal dimension. Section~\ref{sec:text} discusses how data redundancy in the text is handled. Most of the work is based on Transformer models. In both Sections~\ref{sec:video} and~\ref{sec:text}, the primary dimension we use to group the various explorations is how data redundancy is exploited.
\section{Leverage Redundancy in Image Data} \label{sec:image}
\afterpage{
\begin{longtable}[]{|p{3cm}|l|l|p{4.5cm}|p{4.5cm}|}
\caption{Summary of works on image data redundancy}
\label{tab:image}
\\ \hline
\textbf{Application} & \textbf{Paper} & \textbf{Code} & \textbf{Key Idea} & \textbf{Performance} 
\\ \hline
\endfirsthead
\multicolumn{5}{c}
{\tablename\ \thetable\ -- \textit{Continued from previous page}} \\ \hline
\textbf{Application} & \textbf{Paper} & \textbf{Code} & \textbf{Key Idea} & \textbf{Performance} 
\\ \hline
\endhead
image classification & \cite{ning2019deep, deepreuseicde} & N/A                                   & \textit{Deep Reuse} clusters the sub-vectors in activation maps by similarity and reuses the cluster centroids' computation results                                             & $2.41\times$-$3.64\times$ speedup for AlexNet and $2.26\times$-$3.35\times$ for VGG-16 on GPU  
\\ \cline{2-5} 
                                                    ~ & \cite{georgiadis2019accelerating}  & N/A                                   & Sparsify the activation maps                                                                                    & $2.32\times$ speedup on LeNet-5, $1.61\times$ on MobileNet-v1, $2.12\times$ on Inception-v3, $1.77\times$ on ResNet-18, and $1.94\times$ on ResNet-34 
\\ \cline{2-5} 
                                                    ~ & \cite{park2019accelerating}        & N/A                                   & Use a feature map cache to reuse similar feature maps' computation results                                                                              & Evaluated on AlexNet, GoogLeNet, VGG-16, VGG-19, ResNet-50, and ResNet-101, giving $1.06\times$-$1.34\times$ average speedup 
\\ \cline{2-5} 
                                                    ~ & \cite{figurnov2016perforatedcnns}  & \cite{figurnov2016perforatedcnnslink} & \textit{Perforated CNNs} applies perforation masks to skip pixels or patches on activation maps                                                                  & AlexNet with $2\times$-$3.6\times$ speedup on CPU and $2\times$-$4\times$ on GPU; VGG-16 with $2\times$-$4\times$ speedup on CPU and $2\times$-$4\times$ on GPU; NIN~\cite{lin2013network} with $2.2\times$-$4.2\times$ speedup on CPU and $2.1\times$-$3.5\times$ on GPU
\\ \cline{2-5} 
                                                    ~ & \cite{akhlaghi2018snapea}          & N/A                                   & \textit{SnaPEA} predicts the signs of activation outputs earlier to skip potential zero values that appear after the ReLU activations                                          & $28\%$ speedup and $16\%$ energy reduction with no accuracy loss on CNNs (AlexNet, GoogleLeNet, SqueezeNet, VGGNet),  without accuracy loss. $2.02\times$-$3.59\times$ speedup and $1.89\times$-$3.14\times$ energy reduction with $3\%$ accuracy loss 
\\ \cline{2-5} 
                                                    ~ & \cite{hu2016network}               & \cite{hu2016networklink}              & Skip some neurons' computation based on Average Percentage of Zeros (APoZ)                                                                              & $2\times$-$3\times$ less parameters with no accuracy loss on LeNet and VGG-16    
\\ \cline{2-5} 
                                                    ~ & \cite{shomron2019thanks}           & \cite{shomron2019thankslink}          & Zero activation predictor (ZAP) to dynamically predict the zero-valued outputs in the activation maps                                                    & $38\%$-$51\%$ MAC reduction on AlexNet, $26.2\%$-$44.5\%$ on VGG-16, and $23.8\%$-$34\%$ on ResNet-18 on GPU (Nvidia Titan V)
\\ \cline{2-5} 
                                                    ~ & \cite{chen2019layer}               & N/A                                   & Layer-by-layer neuron pruning                                                                                                               & $5.13\times$ speedup on VGG-16 and $3\times$ speedup on ResNet-50 on GPU (Nvidia Titan X (Pascal))
                                                    \\ \cline{2-5} 
                                                     & \cite{suzuki2018spectral}          & N/A                                   & A spectral pruning method that compresses the CNNs by using their \textit{degrees of freedom}                                                          & $32\%$-$55\%$ FLOPs reduction on ResNet-50 
                                                    \\ \cline{2-5} 
                                                     & \cite{gao2018channels}             & \cite{gao2018channelslink}            & Use feature boosting and suppression (FBS) to emphasize salient convolutional channels and skip unimportant ones                                               & $5\times$ speedup on VGG-16 and $2\times$ speedup on ResNet-18
                                                    \\ \cline{2-5} 
                                                     & \cite{ibrokhimov2020effective}     & N/A                                   & Select neurons by the magnitude of their average activations                                                                                            & $2\times$-$3\times$ training speedup by employing $6\times$-$10\times$ fewer parameters on LeNet, AlexNet, and VGG-16 on GPU (Nvidia Titan X (Pascal))
                                                    \\ \cline{2-5} 
                                                     & \cite{lee2020subflow}              & \cite{lee2020subflowlink}             & Apply ranking on the neurons of DNNs based on their contribution to the inference accuracy                                                              & $1\times$-$6.7\times$ speedup on LeNet-5, AlexNet, and KWS \cite{sainath2015convolutional} on GPU (Nividia RTX 2080Ti (Turing) and Jetson Nano)/CPU (x86 and ARM)
                                                    \\ \cline{2-5} 
                                                     & \cite{huang2018data}               & \cite{huang2018datalink}              & Propose to add a sparsity regularization on neurons that guides the selection of more informative neurons                                                          & ResNet-38 achieves $14\%$ less FLOPs and $0.2\%$ lower top-5 error than DenseNet-121 on GPU
                                                    
                                                    \\ \cline{2-5} 
                                                     & \cite{chitsaz2020acceleration}     & N/A                                   & Propose splitting and overlap-and-add operations for FFT-based CNNs to eliminate computation redundancy on activation maps                                         & VGG-16 with single channel input achieve over $3\times$ speedup on FPGA (Xilinx XC6VLX240T)
                                                    \\ \hline
medical image segmentation (3D-UNet)                 & \cite{chen2020frequency}           & N/A                                   & \textit{Frequency Domain Compact 3D CNNs (FDC3D)} eliminating redundancy of 3D convolution filters and feature maps by converting them into the frequency domain               & $2\times$ speedup ratio on 3D-ResNet-18 
                                                    \\ \hline
object detection/recognition                         & \cite{gao2018dynamic}              & N/A                                   & Use two networks to focus on the region of interest on the input data                                                                                               & reducing the number of processed pixels by ~$70\%$ and the detection time by over $50\%$ on pedestrian detection on GPU
                                                    \\ \cline{2-5} 
                                                     & \cite{li2019zoom}                  & \cite{li2019zoomlink}                 & The author implements a zoom network with \textit{map attention decision (MAD) unit} that uses a selective vector to decide the attention on neurons                            & $78.1\%$ mAP and $79.8\%$ mAP compared to SSD baseline on PASCAL with $76.8\%$ mAP
                                                    \\ \cline{2-5} 
                                                     & \cite{de2019skipping}              & N/A                                   & A software-based memoization technique that groups CNN's layer output into range-based clusters and performs table lookup for similar computations          & $3.5\times$ speedup with YOLOv3-tiny and $22\%$ energy consumption reduction on CPU (Intel Core i5-7500, 4 cores, 3.40GHz)
                                                    \\ \hline
image classification \& object detection/recognition & \cite{chen2020vip}                 & \cite{chen2020viplink}                & \textit{ViP} leverages virtual pooling that applies larger strides in the convolution layer to reduce computation                                                              & VGG-16 with $2.1\times$ ($5.23\times$ combined model compression); Faster-RCNN with $1.8\times$ on GPU (Nvidia Titan X and Jetson TX1)                                       
                                                    \\ \cline{2-5} 
                                                     & \cite{chen2019muffnet}             & N/A                                   & \textit{Multi-LayerFeature Federation Network (MuffNet)} splits the activation maps of a layer into several groups and shares the group information with other layers only once & WORK ON small models (under 45 MFLOPs); $0.3\%$ better accuracy than MobileNet with half of its cost.
                                                    \\ \hline
image classification \& video action recognition     & \cite{chen2019drop}                & \cite{chen2019droplink}               & \textit{OctaveConv} factorizes the activation maps into high and low-frequency groups and performs low-cost information exchange and update                              & $82.9\%$ top-1 classification accuracy by ResNet-152 with 22.2 GFLOPs on GPU (Nvidia V100)
                                                    \\ \hline
\end{longtable}
}%

Image data are generally presented as three-dimensional pixel values (height, width, and channel). The inherent locality within data values provides opportunities for advanced optimizations. This section groups the relevant studies at a high level based on the types of redundancy they mainly target: \textit{repeated information}, \textit{irrelevant information}, and \textit{over-detailed information}. The discussion in each kind then divides the techniques to detect and handle data redundancy. Table~\ref{tab:image} presents these works by their applications (tasks) and reported performance for fast reference. The second column (“Paper”) shows the references to the relevant papers; the third column (“Code”) shows the link to the source code of the implementation; the fourth column ("Key Idea") briefly summarizes the key idea of each work. The fifth column (“Performance”) summarizes the performance gains by the data redundant optimization techniques.

\subsection{Repeated Information}
Due to the common coherence of data values in an image, image data often show a certain degree of data locality along the spatial dimensions. The most common optimizations take advantage of the value similarities among pixels. The initial motivation is to minimize the storage space of CNN's intermediate activation maps on a resource-constrained platform and reduce the data movements between CPU and GPU and between processing units and memory. The optimizations also reduce the number of computations and improve computation speeds. 

Table (or cache) lookup~\cite{park2019accelerating, de2019skipping, mocerino2019energy, razlighi2017looknn, ma2020axr, hegde2018ucnn} and clustering~\cite{kim2020power, ning2019deep} are the most typical approaches taken to discover and leverage repeated information in images.

To detect similarity among data points, the various techniques differ in the level of pixel units used for the similarity detection, some at the level of bits~\cite{jiao2018energy, ma2020axr, salamat2018rnsnet}, some at the level of pixels~\cite{razlighi2017looknn}, patches (or tiles, sub-vectors)~\cite{kim2020power, ning2019deep, hegde2018ucnn}, or even entire feature maps as a whole~\cite{park2019accelerating, de2019skipping, mocerino2019energy, jiao2018energy}.  Some use clustering with predefined distance metrics and thresholds to identify the similarity between units, while others use hash functions like locality sensitive hashing~\cite{ning2019deep} or bloom filters~\cite{jiao2018energy}. RNSnet~\cite{salamat2018rnsnet} utilizes a unique bit-handling technique that maps the binary representation into the Residue Number System (RNS). Each binary representation is divided by a modulus set; with the values represented by the corresponding modules and reminders, the technique efficiently identifies similar values. The method could be regarded as a kind of hashing function. 

After gathering similar data points, each bucket (or cluster) of the data is represented by usually one or a few data points in the bucket. The determination of such representatives depends on the tasks (e.g., image classification, object detection) that the DNN performs. For a relatively simple image classification task, if the number of buckets is enough to showcase the overall discrepancy between predicted classes without hurting the training or inference accuracy, the representative can be simply an average of data values in that bucket. To the extreme, when the dataset contains a significant proportion of identical data samples, a representative is sometimes set as an arbitrary data point taken from a bucket. The representative is used in place of the other issues in the bucket during the CNN inference or training, such that similar computations can be avoided. 

\begin{figure}[ht]
  \includegraphics[width=0.6\textwidth]{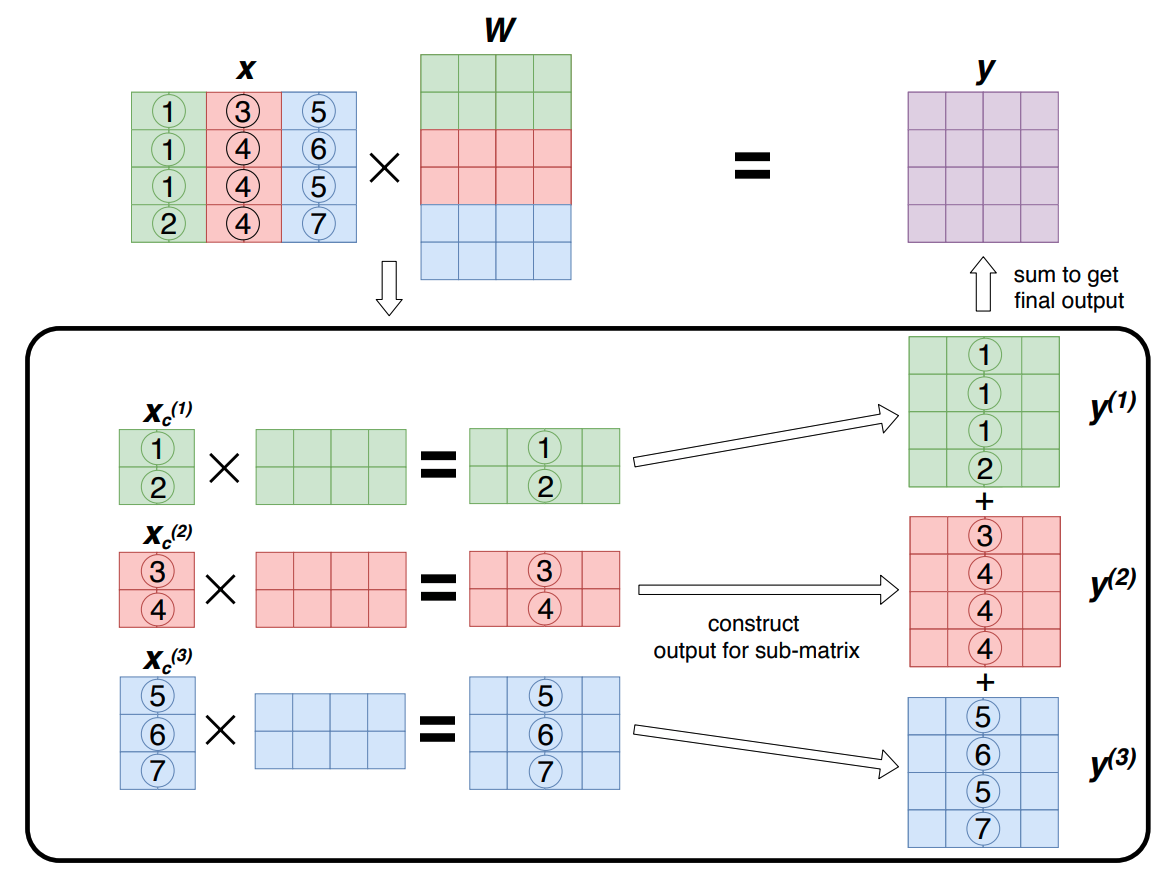}
  \caption{Illustration of sub-vector clustering in \textit{Deep Reuse}. In a convolution layer, the activation maps of 2D images are unfolded as $x$ and are multiplied by weights $W$ of corresponding filters. With sub-vector clustering, $x$ is first divided into sub-vectors and labeled with corresponding cluster ids. In the matrix multiplication, only the representative of each cluster is used. Afterward, the output activation maps are re-constructed with the representative results of each cluster. Figure adapted from reference~\cite{ning2019deep}.}
  \label{fig:deepreuse}
\end{figure}

Representative work in this category is the Deep Reuse work by Ning and others~\cite{ning2019deep, deepreuseicde}. As shown in Fig.~\ref{fig:deepreuse}, the approach divides input images or activation maps into sub-vectors during runtime. Based on their similarities, it clusters them into several buckets through efficient online LSH-based clustering~\cite{10.5555/645925.671516}, with each cluster represented by its centroid. Besides handling repeated information in a 2D scope, Deep Reuse also addresses the redundancy in data batches. Based on the clustering results, Deep Reuse reduces a convolution to multiplications between a small matrix formed by the clusters centroids and the filter matrix. The results are used to reconstruct the full activation map through data duplications. Even with the online clustering overhead, the work shows that the reuse of computation results yields an overall $1.77-2\times$ speedup (up to $4.3\times$ layer-wise) with negligible accuracy loss and without model retraining. The method yields more speedups when applied to CNN training in an adaptive manner~\cite{deepreuseicde}.

A fundamental tradeoff facing the methods in this category is the accuracy and the amount of computation reuse. It is determined by the aggressiveness in similarity-based clustering. For instance, in the Deep Reuse work, the aggressiveness is determined by two hyper-parameters, the length of a sub-vector in an activation map (granularity), and then the number of LSH hashing vectors. As the appropriate values of the hyperparameters are data and model dependent, the previous work~\cite{ning2019deep} employs some offline tuning process to select the values such that the accuracy is not compromised. One of the directions worth some future explorations is to adapt the clustering method to data and models. For instance, in Deep Reuse, the hashing vectors are randomly generated. If they can be learned from data for a given CNN, more aggressive clustering might be possible without causing accuracy losses.

\subsection{Irrelevant Information}
\subsubsection{Feature Pruning}
Image data or intermediate activation maps may have unnecessary information as well. For example, most well-trained CNNs can still maintain their prediction accuracy when some pixels in the activation maps are removed. Subsequently, the skipped portion of the data reduces the amount of computation and thus offers acceleration during inference or training. We introduce some details about how the optimizations in this line identify these irrelevant values at different scales and get rid of them while keeping the overall accuracy.

Hu et al.~\cite{hu2016network} identify many zeros in the activation of CNNs. They quantify the neurons to be pruned based on the \textit{Average Percentage of Zeros (APoZ)}. \textit{APoZ} is calculated on a per-layer basis to measure the probability each fixed position appears to be zero-valued in validation data samples. With the same motivation, Akhlaghi et al.~\cite{akhlaghi2018snapea} and Piyasena et al.~\cite{piyasena2019reducing} propose detecting zero pixels that may occur after the ReLU activations. It is by identifying the negative activations earlier using a low-cost approximation scheme since the negative values turn into zeros after going through the ReLU layers. In particular, Akhlaghi et al.~\cite{akhlaghi2018snapea} construct \textit{SnaPEA} with a runtime technique to speculate on the convolution outputs' sign before going through negative weights. The aggressiveness of the speculation is controlled by thresholding on parameters and the predefined associated number of MAC operations that can maintain accuracy. The parameters are tuned offline within the search space. Piyasena et al.~\cite{piyasena2019reducing} augment the CNN implementation with a lightweight approximation scheme that consists of \textit{ApproxConv} and \textit{ReLupred} stages. The two operations work sequentially to handle the sign prediction on an approximate convolution computation.

\begin{figure}[ht]
  \includegraphics[width=0.75\textwidth,height=0.75\textheight,keepaspectratio,trim={0 5.5cm 0 3cm},clip]{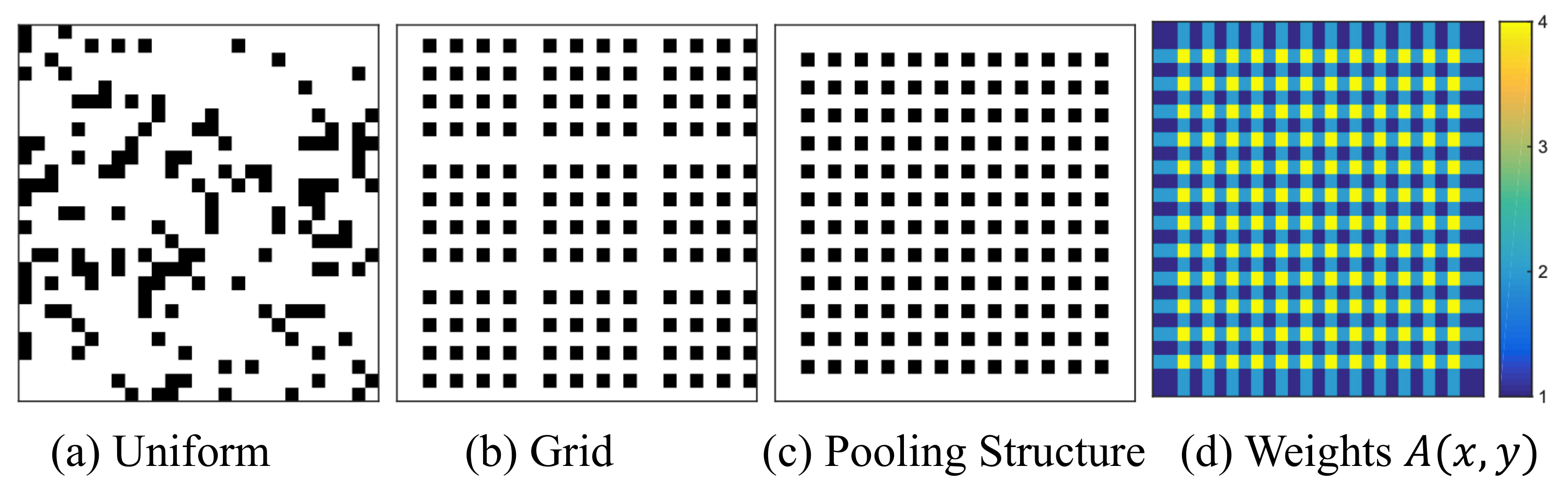}
  \caption{Input-independent perforation masks in \textit{PerforatedCNNs}. According to the paper, the mass layout is for the AlexNet Conv2 layer. (a) \textbf{Uniform perforation mask} is $N$ points chosen randomly without replacement from the set $\Omega$. (b) \textbf{Grid perforation mask} is a set of points \begin{math} I = {a(1), . . . , a(Kx)} \times {b(1), . . . , b(Ky)} \end{math} the values of a(i), b(i) using the pseudorandom integer sequence generation scheme~\cite{graham2014fractional}. (c) \textbf{Pooling structure mask} exploits the structure of the overlaps of pooling operators. Denote by $A(x, y)$ the number of times the convolutional layer output is used in the pooling operators. The pooling structure mask is obtained by picking top-N positions with the highest values of $A(x, y)$. (d) \textbf{Impact mask} estimates the impact of perforation of each position on the CNN loss function and then removes the least important positions. When activation maps are computed, the coordinates of black pixels are skipped. Figures and description adapted from reference~\cite{figurnov2016perforatedcnns}.}
  \label{fig:perforatedcnns}
\end{figure}

A more direct way to remove irrelevant information is to assume the existence of redundancy based on domain knowledge and prune the data or activation maps at different scales. Figurnov et al.~\cite{figurnov2016perforatedcnns} prunes the data at pixel and patch level. The proposed mechanism inherits the idea of the loop perforation technique in code optimization to skip certain spatial positions in images for CNN classification inference. The selection of pixels to prune is random within a predefined limited number of points. They exploit four input-independent perforation masks: \textit{Uniform mask}, \textit{Grid mask}, \textit{Pooling Structure mask}, and \textit{Impact mask}, as shown in Fig.~\ref{fig:perforatedcnns}. These masks are used to confine the random point selected into corresponding patterns. Afterward, perforated CNN uses interpolation to reconstruct the output feature maps. The CNN wights are retrained to adapt to this optimization as well.


Aside from the pruning technique within the scope of a single activation map, irrelevant information can be addressed across the channel dimensions by, for instance, removing a set of outside 2D feature maps. Specifically, Gao et al.~\cite{gao2018channels} apply feature boosting and suppression by formulating a channel saliency predictor. The predictor serves as an indicator to skip execution at some feature maps at runtime. Chakraborty et al.~\cite{chakraborty2019feature} show that in a simple handwritten digit recognition task, directly discarding redundant feature maps randomly before the full connection does not affect the prediction accuracy much. On the same axis, Singh et al.~\cite{singh2019deep} conduct a comprehensive analysis on pruning redundant activation maps in 1D CNN, SoundNet~\cite{aytar2016soundnet}, at a per map level. The pruned feature map selection is based on the following measurements: ANOVA-based method~\cite{penny2006analysis}, Entropy-based (DE) method~\cite{perez2009estimation}, Cosine-similarity (CS) based method, and a greedy algorithm for selection of feature maps based on KL divergence. Gaikwad et al.~\cite{gaikwad2019pruning} apply the L2 norm to decide feature map importance in pruning feature maps of SqueezeNet~\cite{iandola2016squeezenet}. Although the technique is not for image data, the selection criteria are worthy of being noted under the context of irrelevancy elimination. Furthermore, multiple granularities of redundancy can be handled at the same time. For example, Yu et al.~\cite{yu2020antidote} combine spatial and channel-wise neuron pruning on activation maps. The criteria of what to be pruned are based on the attention mechanism. The attention mechanism is trained with targeted dropout, enhancing inference time robustness without accuracy loss after pruning. 

Moreover, the optimizations can be classified by whether the irrelevancy is subject to the input datasets or not. Namely, they are input-dependent or input-independent. Input-dependent irrelevant information removal can be achieved by detecting data irrelevancy based on statistical~\cite{hu2016network, singh2019deep, gao2018channels} or learning-based heuristic~\cite{akhlaghi2018snapea, yu2020antidote}, or estimation of values~\cite{piyasena2019reducing}. On the contrary, input-independent approaches~\cite{figurnov2016perforatedcnns, chakraborty2019feature} get rid of those irrelevant values by implicitly assuming them to appear randomly with a limit threshold in structural or non-structural manners.

\subsubsection{Neuron or Feature Subset Selection}
Another granularity used in feature map pruning is the selection of neurons (or activation). It could be regarded as a feature selection process at the value granularity. The neurons not chosen can be interpreted as ones being pruned. A subset of activations is determined based on specific criteria in neuron selection. Lee et al.~\cite{lee2020subflow} apply ranking on the neurons of DNNs based on their contribution to the inference accuracy. Huang et al.~\cite{huang2018data} add a sparsity regularization on neurons to force some of them to become zeros and guide the selection of more informative neurons without further tuning. Besides, Ibrokhimov et al.~\cite{ibrokhimov2020effective} propose the technique that selects neurons based on the magnitude of their average activations and keeps only the exact amount of the most minor activated neurons that work well. They report a reduction in the total number of operations and corresponding speedups. 

\subsubsection{Dynamic Region of Interest}
\begin{figure}[ht]
  \includegraphics[width=\textwidth,height=\textheight,keepaspectratio]{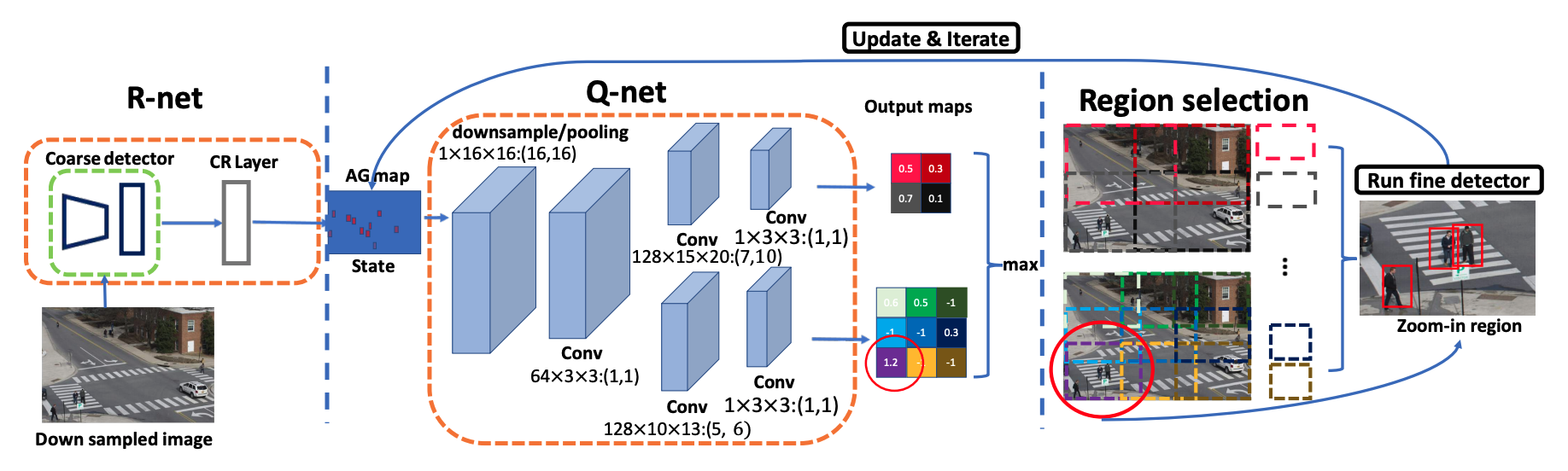}
  \caption{The workflow of \textit{R-net} and \textit{Q-net}. Given a down-sampled image as input, the R-net generates an initial accuracy gain (AG) map indicating the potential zoom-in accuracy gain of different regions (initial state). The Q-net is applied iteratively on the AG map to select regions. Once a region is selected, the AG map will be updated to reflect the history of actions. Two parallel pipelines are used for the Q-net, each of which outputs an action-reward map that corresponds to selecting zoom-in regions with a specific size. The map's value indicates the likelihood that the action will increase accuracy at a low cost. Action rewards from all maps are considered to select the optimal zoom-in region at each iteration. The notation $128 \times 15 \times 20:(7,10)$ means 128 convolution kernels with size $15 \times 20$, and stride of 7/10 in height/width. Each grid cell in the output maps is given a unique color, and a bounding box of the same color is drawn on the image to denote the corresponding zoom region size and location. Figure and description adapted from reference~\cite{gao2018dynamic}.}
  \label{fig:zoomin}
\end{figure}

In object detection, dynamic zooming in~\cite{gao2018dynamic, li2019zoom} of a particular region of interest can help reduce computations on irrelevant background information. Gao et al.~\cite{gao2018dynamic} present such a technique. They employ \textit{zoom-in accuracy gain regression network (R-net)} and \textit{zoom-in Q function network (Q-net)}. The \textit{R-net} learns the correlation between coarse and fine detection and predicts the accuracy gain for zooming in to a specific region. The \textit{Q-net} learns via reinforcement learning. It sequentially selects the optimal zoom locations and scales them correspondingly by analyzing the new output of \textit{R-net} and its history. In the inference phase, a down-sampled image is input to the \textit{R-net} and predicts the accuracy gain. The gain serves as an indicator for the \textit{Q-net} to select the target on regions to concentrate sequentially. The workflow is illustrated in Fig.~\ref{fig:zoomin}. The mechanism reduces the total pixels in computation by over $50\%$ and reduces the average detection time by $25\%$ on the Caltech Pedestrian Detection dataset. It also reduces the pixels by $70\%$ and obtains over $50\%$ detection time reduction on the YFCC100M dataset. 

\subsection{Over-detailed Information}
\begin{figure}[ht]
  \includegraphics[width=0.7\textwidth,height=0.7\textheight,keepaspectratio, trim={5cm 0 3cm 0},clip]{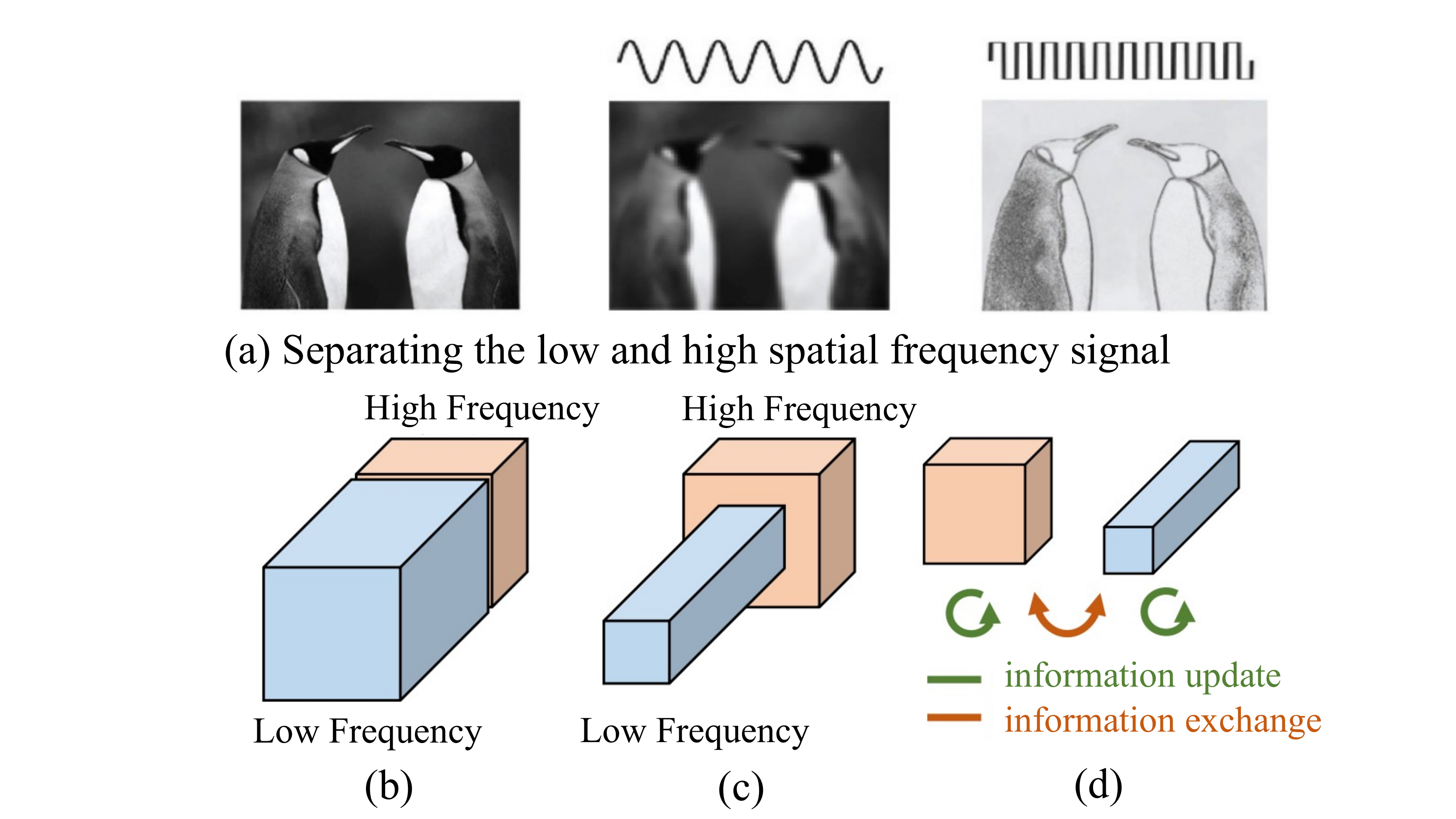}
  \caption{Illustration of \textit{Octave Conv} in the factorization of high and low spatial frequency. (a) Motivation. The spatial frequency model for vision~\cite{campbell1968application, de1980spatial} shows that raw images can be decomposed into a low and a high spatial frequency part. (b) The output maps of a convolution layer can also be factorized and grouped by their spatial frequency. (c) The proposed multifrequency feature representation stores the smoothly changing, low-frequency maps in a low-resolution tensor to reduce spatial redundancy. (d) The proposed Octave Convolution operates directly on this representation. It updates the information for each group and further enables information exchange between groups. Figures and descriptions adapted from reference~\cite{chen2019drop}.}
  \label{fig:octaveconv}
\end{figure}

Yet another direction in image data redundancy exploitation is eliminating over-detailed information, represented by adaption to variant image resolutions, especially resolution lowering. Chen et al.~\cite{chen2019drop} propose Octave convolution (\textit{OctConv}) (Fig.~\ref{fig:octaveconv}) to leverage the spatial redundancy and factorize the activation maps into high and low frequency (resolution) groups to save both memory and computation cost. The operations at each layer enable the information to be exchanged within both groups during computation. It is implemented as a portable plug-and-play convolution unit and can be applied together with a compressed model or optimization that reduces channel-wise redundancy. An OctConv-equipped ResNet-152 can achieve $82.9\%$ top-1 classification accuracy on ImageNet with $22.2$ GFLOPs. 

\begin{figure}[ht]
  \includegraphics[width=0.75\textwidth,height=0.75\textheight,keepaspectratio]{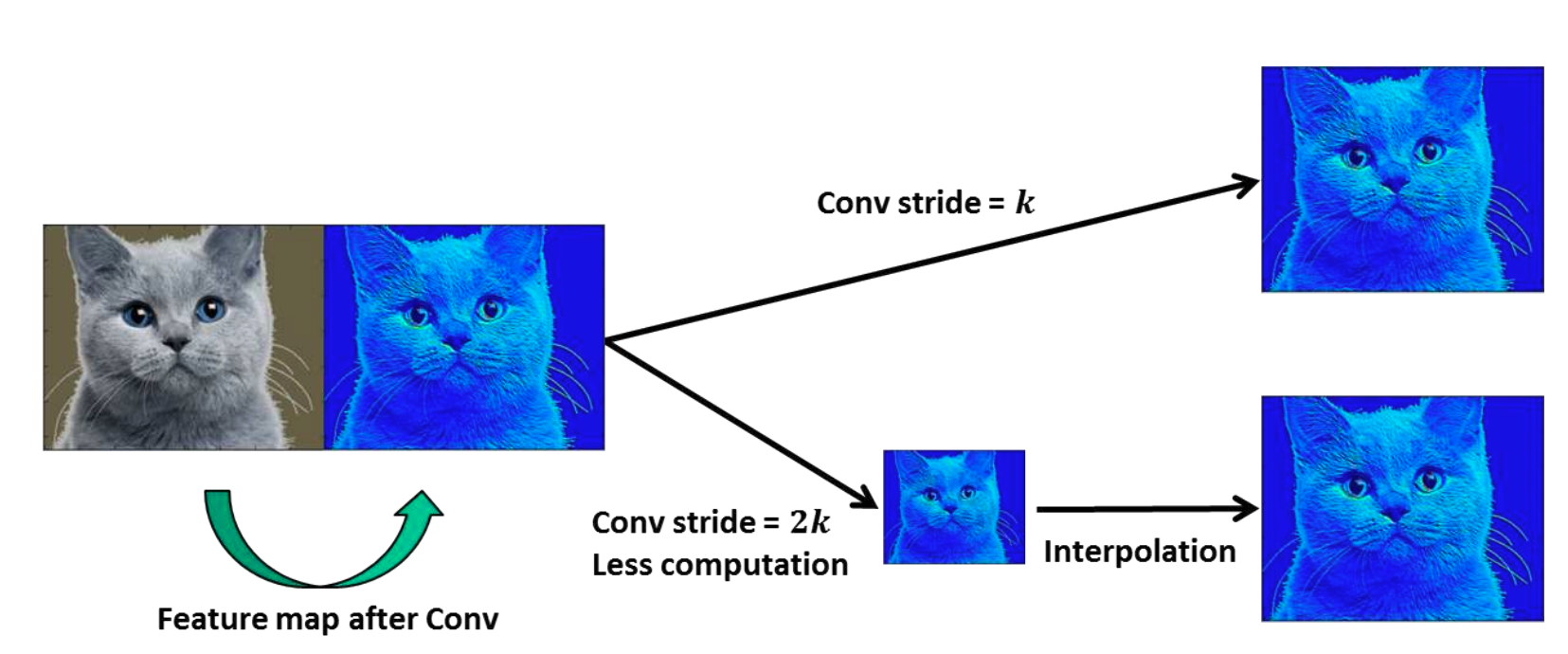}
  \caption{Illustration of virtual polling~\cite{LinkViz}. By using a larger stride, it save computation in conv layers and, to recover the output size, it use linear interpolation which is fast to compute. Figure and description adapted from reference~\cite{chen2020vip}.}
  \label{fig:vip}
\end{figure}

\textit{ViP} proposed by Chen et al.~\cite{chen2020vip} also targets the spatial redundancy in feature maps. It takes advantage of virtual pooling, as shown in Fig.~\ref{fig:vip}, such that the larger stride is used to obtain a smaller feature map and at the same time save convolution computation. To recover the output feature map dimension, linear interpolation is applied. ViP delivers $2.1\times$ speedup with less than $1.5\%$ accuracy degradation in ImageNet classification on VGG16, and $1.8\times$ speedup with $0.025$ mAP degradation in PASCAL VOC object detection with Faster-RCNN. ViP also reduces mobile GPU and CPU energy consumption by up to $55\%$ and $70\%$, respectively. 

\section{Leverage Data Redundancy in DNN for Videos} \label{sec:video}

Existing DNN frameworks for video analysis are built on well-developed CNNs such as VGG~\cite{simonyan2014deep} or ResNet~\cite{he2015deep} and image-based object detection algorithms such as Faster-RCNN~\cite{ren2015faster}. They treat the video as a continuous input of images. Intuitively, the data redundancy in this line of works, especially the duplication along the temporal axis, provides much room for performance enhancement. DNN frameworks that leverage data redundancy in video data can be categorized based on how they exploit redundancy. These techniques fall into the three categories listed in our taxonomy (skipping, reuse, and approximation); this section groups them more detailed to capture the more complicated differences among the techniques. Table~\ref{tab:video} shows the works with video data redundancy exploitation by their applications.

\afterpage{
\begin{longtable}[]{|p{3cm}|l|l|p{4.5cm}|p{4.5cm}|}
\caption{Summary of works on video data redundancy}
\label{tab:video}
\\ \hline
\textbf{Application} & \textbf{Paper} & \textbf{Code} & \textbf{Key Idea} & \textbf{Performance} 
\\ \hline
\endfirsthead
\multicolumn{5}{c}%
{\tablename\ \thetable\ -- \textit{Continued from previous page}} \\
\hline
\textbf{Application} & \textbf{Paper} & \textbf{Code} & \textbf{Key Idea} & \textbf{Performance} 
\\ \hline
\endhead
surveillance video processing                                        & \cite{cavigelli2017cbinfer} & \cite{cavigelli2017cbinferlink} & \textit{CBInfer} utilizes change-based evaluation for CNNs to detect redundant pixels on video tasks                                                             & Evaluated on a self-constructed 5-layer CNN with input image size $871\times541$, provides speedup $8.6\times$ over the CUDA baseline. Energy efficiency $10\times$ higher (328 GOp/s/W on Nvidia Jetson TX1)                                     \\ \cline{2-3}
                                                                     & \cite{cavigelli2018cbinfer} & \cite{cavigelli2018cbinferlink} &                                                                                                                                                                  &                                                                                                                                                                                                                                     \\ \hline
video object detection                                               & \cite{chin2019adascale}     & N/A                             & \textit{AdaScale} trains a object detector and a regressor to dynamically adjust the optimal detector scale for video frames                                     & Improve the speed by $25\%$; RFCN + AdaS ~20 fps; SeqNMS~\cite{han2016seq} + AdaS ~16 fps; DFF~\cite{zhu2017flowguided} + AdaS ~60 fps on GPU (Nvidia GTX 1080Ti)
                                                        \\ \cline{2-5} 
                                                                     & \cite{chen2018optimizing}   & N/A                             & \textit{Scale-Time Lattice} performs expensive detection with fewer times and propagates the results across scale and time to balance the performance and accuracy tradeoff                                                                                                                                      & On ImageNet VID, the approach achieves $79.6\%$ at 20 fps or $79.0\%$ at 62 fps on GPU                                                                                                                                               \\ \cline{2-5} 
                                                                     & \cite{liu2017mobile}        & \cite{liu2017mobilelink}        & A convolutional LSTM to assist Single Shot Detector (SSD) on refining feature maps and reuse history information                             & 15 fps with MobileNet-SSD on CPU                                                                                                                                                                                                    \\ \hline
video object tracking                                                & \cite{mao2018catdet}        & N/A                             & \textit{CaTDeT} applies a tracker to assist refining intermediate feature maps for better reuse                                                                  & Constructed Res10a and Res10b as proposal net, ResNet50 as refinement net, giving $81.4\%$ and $81.5\%$ mAP with MAC size 49.3 Gops and 29.3 Gops respectively on GPU (Nvidia Titan X (Maxwell)) 
                                                                    \\ \cline{2-5} 
                                                                     & \cite{zhang2017kill}        & N/A                             & \textit{Kill Two Bird with One Stone} aggregates moving objects as patch-of-interest to construct a compact patch that skips duplicate image patchs' calculation & SSD with $300\times300$ subframe size at 13.7-24.2 fps and $500\times500$ subframe size at 23.8-25.7 fps on GPU (Nvidia Titan X)
                                                    \\ \hline
temporal action localization                                         & \cite{yeung2015endtoend}    & \cite{yeung2015endtoendlink}    & An end-to-end model for temporal action localization that generates compact video representations                                           & Requiring $2\%$ or less frames. 
                                                                                        \\ \cline{2-5} 
                                                                     & \cite{alwassel2017action}   & \cite{alwassel2017actionlink}   & \textit{Action Search} uses LSTMs to predict next search location from current frame's location and the history                                              & Observing an average $17.3\%$ of the video $30.8\%$ mAP on human activities recognition 
                                                    \\ \hline
frame selection                                                      & \cite{wu2018adaframe}       & N/A                             & \textit{AdaFrame} uses memory-augmented LSTM to adaptively select relevant frames for fast prediction                                                            & On FCVID~\cite{jiang2017exploiting} and ActivityNet dataset, achieving the same performance with only 8.21 and 8.65 frames, respectively. 
\\ \cline{2-5} 
                                                                     & \cite{korbar2019scsampler}  & N/A                             & \textit{SCSampler} is a lightweight model for clip sampling that invokes recognition on only the most salient clips                                               & Improve the state-of-art 3D CNN video classification models (ResNet3D (R3D), Mixed Convolutional Network(MC3), and R(2+1)D~\cite{du2017closer}) on Sports1M by up to $7\%$ and reduces the computation cost by more than $15\times$ on GPU (32 Nvidia P100)
                                                                     \\ \cline{2-5} 
                                                                     & \cite{kang2017noscope}      & \cite{kang2017noscopelink}      & \textit{NoScope} adopts a difference detector and short-circuit evaluation on the model that allows skipping identical neighboring video frames                  & YOLOv2 with more than 30 fps and less than $5\%$ accuracy loss on GPU (Nvidia P100)
                                                    \\ \hline
action recognition                                                   & \cite{wu2019multiagent}     & N/A                             & The authors use gated recurrent units (GRUs) and policy networks in reinforcement learning to adjust video sampling location per time step                       & $79.1\%$ mAP for ActivityNet-v1.3 on YouTube Birds and $79.77\%$ on YouTube Cars on GPU (Nvidia Titan X)  
                                                    \\ \hline
streaming activity recognition \& untrimmed video activity detection & \cite{su2016leaving}        & N/A                             & An active mechanism that prioritizes frames and makes timely action prediction                                                              & Achieving comparable accuracy on 32 fps and 64 fps CNN video classification on CNNs            
                                                    \\ \hline
video object detection \& video semantic segmentation                & \cite{zhu2016deep}          & \cite{zhu2017highlink}          & \textit{deep feature flow network (DFF)} propagates neighboring frames information with a flow estimation algorithm                                             & 20.25 fps and $73.1\%$ mAP on ResNet-101-R-FCN on GPU (Nvidia K40)                
                                                    \\ \cline{2-2} \cline{4-5} 
                                                                     & \cite{zhu2017flowguided}    &                                 & Assemble the appearance information in consecutive frames                                                                                 & $74.0\%$ mAP and $76.3\%$ mAP on ResNet-50-R-FCN and ResNet-101-R-FCN respectively on GPU (Nvidia K40)
                                                    \\ \cline{2-2} \cline{4-5} 
                                                                     & \cite{zhu2017high}          &                                 & A unified framework that includes the above two techniques                                                                                                       & $77.8\%$ mAP on ResNet-101-DCN on GPU (Nvidia K40) 
                                                    \\ \hline
video object detection \& video object localization                  & \cite{Kang_2017, Kang_2018} & \cite{Kang_2017link}            & Use LSTMs for temporal information propagation in order to improve video detection accuracy                                               & Evaluated on DeepID-Net~\cite{ouyang2014deepidnet} with CRAFT~\cite{yang2016craft} (pretrained Faster-RCNN), offering $67.82\%$ mAP and $68.4\%$ mAP on GPU (4 Nvidia Titan X)    
                                                    \\ \hline

\end{longtable}
}%

\subsection{Pruning}
\begin{figure}[ht]
  \includegraphics[width=0.65\textwidth,height=0.65\textheight,keepaspectratio]{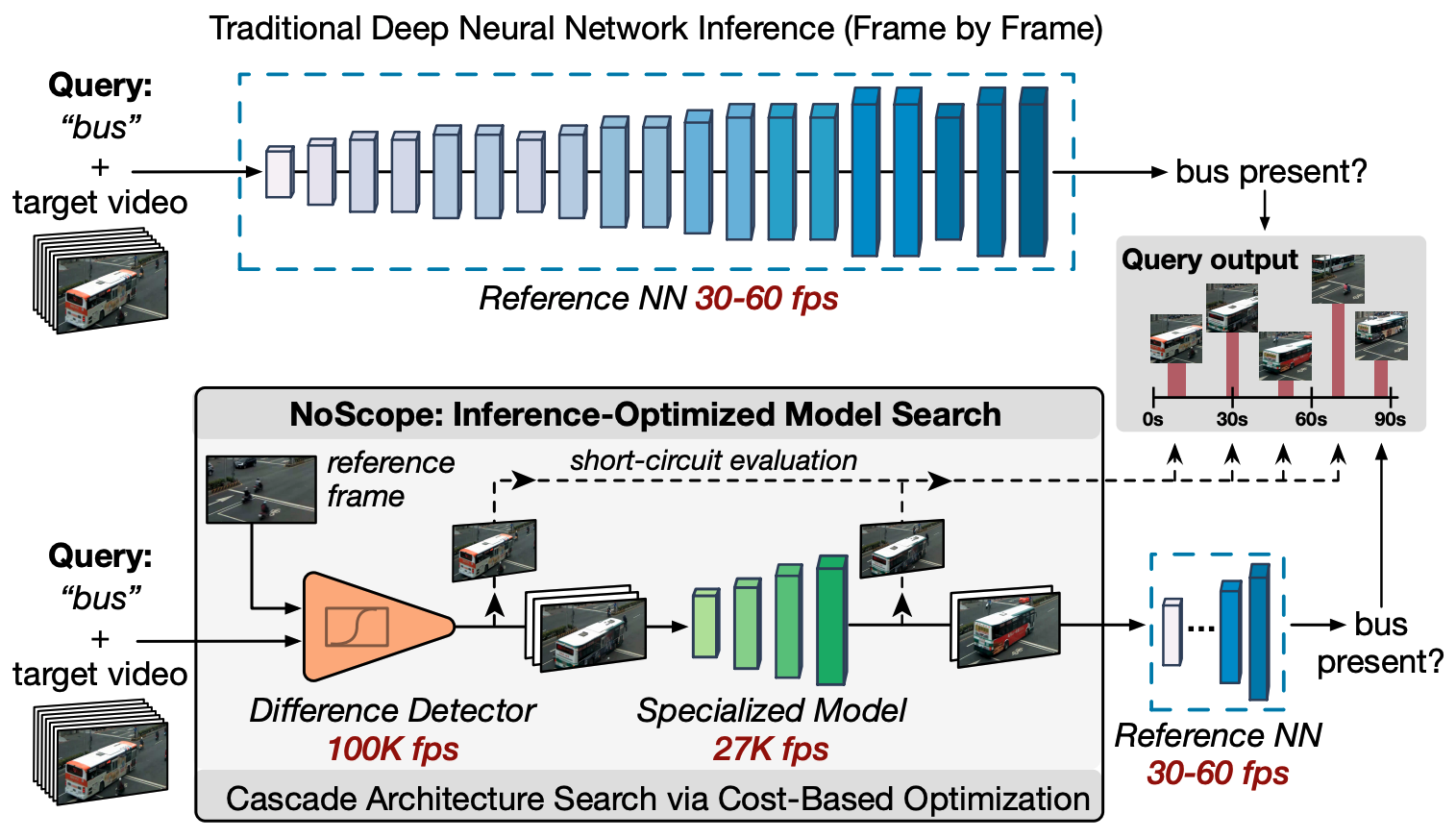}
  \caption{Illustration of the \textit{NoScope} framework. \textit{NoScope} is a system for accelerating neural network analysis over videos via an inference-optimized model search. Given an input video, target object, and reference neural network, \textit{NoScope} automatically searches for and trains a cascade of models—including difference detectors and specialized networks—that can reproduce the binarized outputs of the reference network with high accuracy—but up to three orders of magnitude faster. Figure and description adapted from reference~\cite{kang2017noscope}.}
  \label{fig:noscope}
\end{figure}

A way to avoid recomputing similar values across video frames is through skipping some of them along the temporal dimension (or the batch dimension in 2D CNN). Kang et al. propose \textit{NoScope}~\cite{kang2017noscope} to optimize the processing of video querying. As shown in Fig.~\ref{fig:noscope}, they insert a difference detector and a specialized model with short-circuit evaluation that allows computation reduction on almost identical neighboring frames in a video sequence. 

\begin{figure}[ht]
  \includegraphics[width=\textwidth,height=\textheight,keepaspectratio]{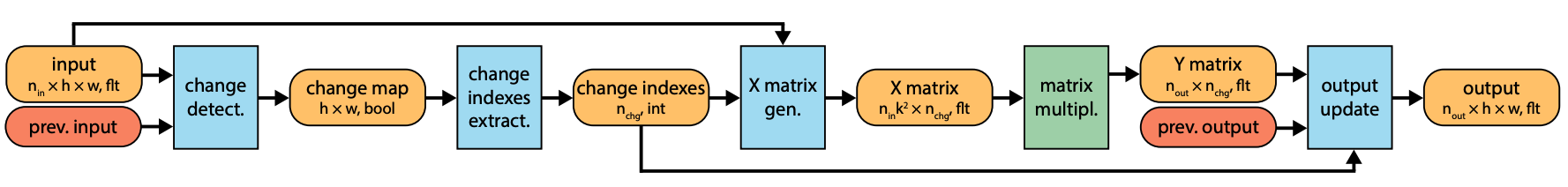}
  \caption{Processing flow of the change-based convolution algorithm. Custom processing kernels are shown in blue, processing steps using available libraries are shown in green, variables sharable among layers are shown in yellow, and variables to be stored per layer are colored orange. The size and data type of the tensor storing the intermediate results is indicated below each variable name. Figure and description adapted from reference~\cite{cavigelli2017cbinfer}.}
  \label{fig:cbinfer}
\end{figure}

The redundancy across continuous frames can also be addressed at the pixel level. Cavigelli et al. propose \textit{CBinfer}~\cite{cavigelli2017cbinfer, cavigelli2018cbinfer}, a change-based evaluation of CNN for video data recorded with a static camera, to exploit the spatial-temporal sparsity of pixel changes. As shown in Fig.~\ref{fig:cbinfer}, they modify each convolution layer to include an additional change detection mechanism and change indexes extraction before the matrix multiplication. The output is updated with both non-changed output pixels from the previous frame and changed output pixels calculated at this time. \textit{CBinfer} achieves an average speed-up of $8.6\times$ over a cuDNN baseline on a realistic benchmark with a negligible accuracy loss of less than $0.1\%$ and no network retraining. The resulting energy efficiency is $10\times$ higher than per-frame evaluation and reaches an equivalent of $328$ $GOp/s/W$ on the Nvidia Tegra X1 platform. 


\subsection{Approximate Representation}
\begin{figure}[ht]
  \includegraphics[width=\textwidth,height=\textheight,keepaspectratio]{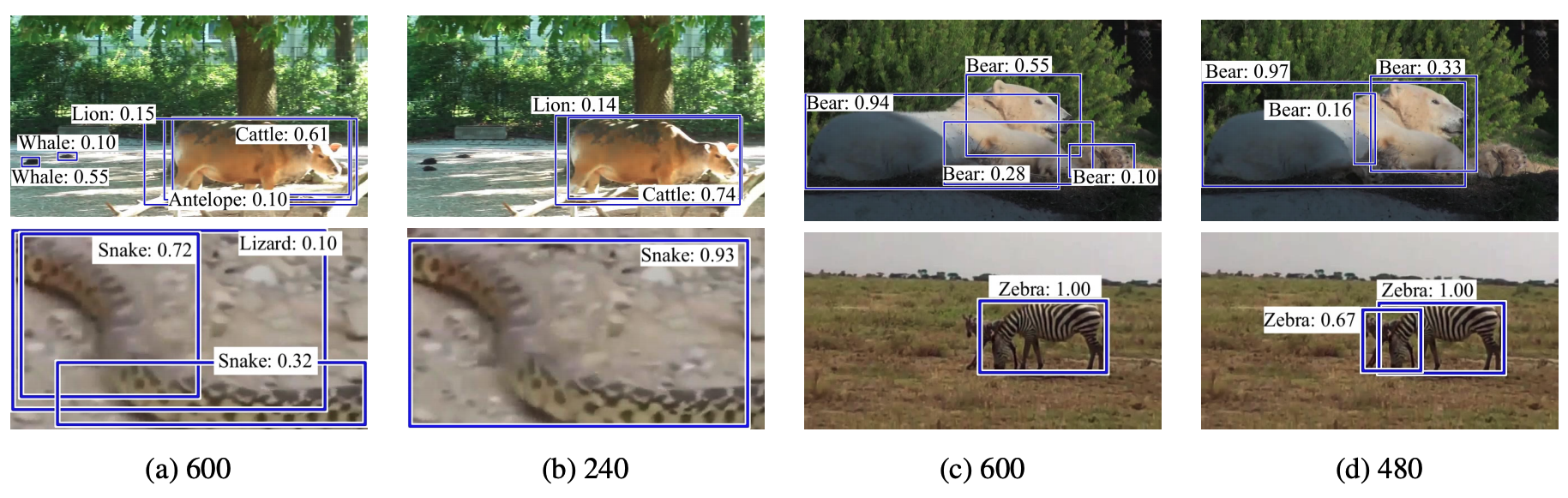}
  \caption{\textit{AdaScale} detection result visualization. Examples where down-sampled images have better detection results. Blue boxes are the detection results, and the numbers are the confidence. The detector is trained on a single scale (pixels of the shortest side) of 600. Column (a) and (c) are tested at scale of 600.
Column (b) is tested at scale 240, and column (d) is tested at scale 480. Figures and descriptions adapted from reference~\cite{chin2019adascale}.}
  \label{fig:adascale}
\end{figure}

Regional re-scaling is a type of approximate representation. Chin et al. propose \textit{AdaScale}~\cite{chin2019adascale} that exploits the potential benefit bought by adaptive image scaling. They resort to image down-sampling to realize performance improvement on both accuracy and speedup for video object detection. To accomplish this, \textit{AdaScale} adopts an object detector to generate the optimal scale labels for images carried out by a learning-based method. The generated optimal scale is later used to train the \textit{scale regressor} that dynamically re-scales the image. The visualization of the effect is shown in Fig.~\ref{fig:adascale}. When the images are down-sampled, they are supposed to reduce the number of false positives introduced by overly detail-oriented messages, which in turn dismiss the redundancy. The image scaling increases the number of true positives by turning the objects too large to smaller ones for the detector to be more confident. For ImageNet VID and mini YouTube-BB datasets, AdaScale demonstrates $1.3\%$ and $2.7\%$ mAP improvement with $1.6\times$ and $1.8\times$ speedup respectively. The framework can also work complementary with video acceleration work~\cite{zhu2016deep}, of which they experience a speedup gain by $25\%$ and a slight mAP boost on the ImageNet VID dataset. 

\subsection{Subset Selection}
Subset selection is for obtaining a subgroup of potentially representative video frames or corresponding feature maps. These techniques adopt a variety of dimensions in preference, from the spatial to the temporal, batch, and streaming.

\subsubsection{Selection in the Spatial Dimension}
\begin{figure}[ht]
  \includegraphics[width=\textwidth,height=\textheight,keepaspectratio]{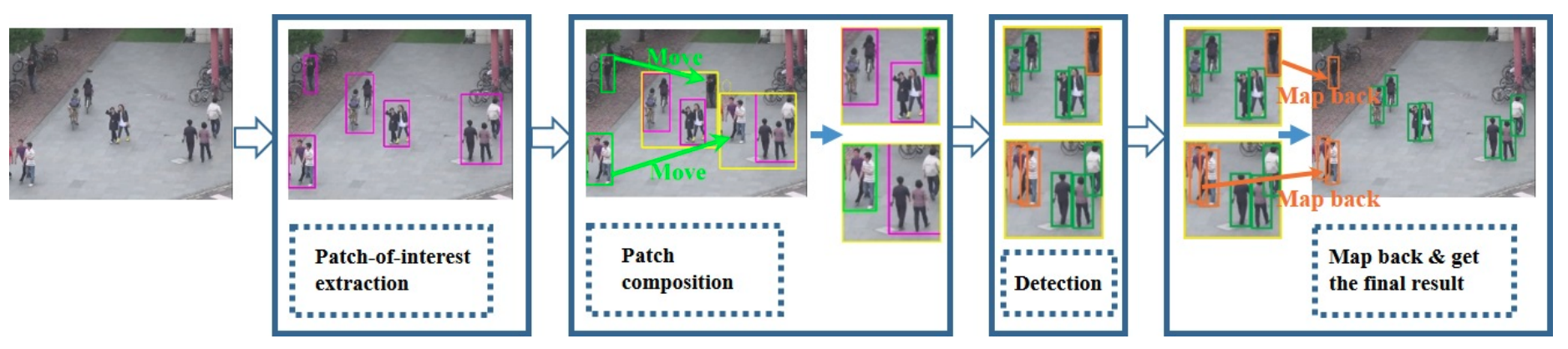}
  \caption{Patch-of-Interest composition. (a) The input image. (b) Detected patches. (c) The patch composition (left) and sub-frames(right). (d) Detection results on sub-frames. (e) Map back and get the final result on the original image. Figures and descriptions adapted from reference~\cite{zhang2017kill}.}
  \label{fig:kill}
\end{figure}

\paragraph{Patch-of-Interest} Zhang et al.~\cite{zhang2017kill} propose \textit{Kill Two Bird with One Stone} to aggregate patch-of-interest (usually the moving object within images) for construction of a compact patch composition for video object detection as shown in Fig.~\ref{fig:kill}. In the detection steps, the parts of the spatial dimension data patches are eliminated to save unnecessary computation. The set of patch-of-interest is mapped back by their numbers and location offsets to reconstruct the final result. 

\begin{figure}[ht]
  \includegraphics[width=0.75\textwidth,height=0.75\textheight,keepaspectratio]{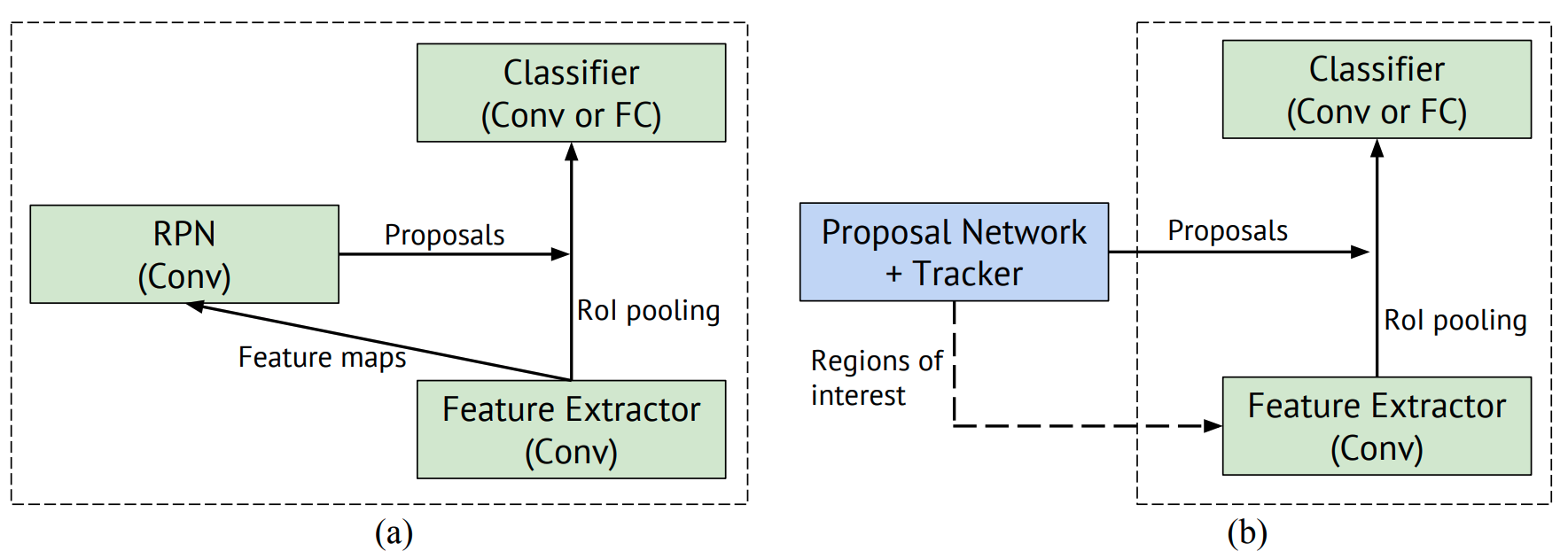}
  \caption{Illustration of algorithm flow in CaTDeT. Compare the inference-time workflows of the standard Faster R-CNN model and the refinement network. (a) Standard Faster R-CNN detector: the RPN takes the feature maps from the feature extractor. (b) Faster R-CNN detector for the refinement network: the proposals from the proposal network and the tracker instruct the feature extractor only to compute features on regions of interest. Regions of interest are a mask of all proposals over the frame. Figures and descriptions adapted from reference~\cite{mao2018catdet}.}
  \label{fig:catdet}
\end{figure}

\paragraph{Patch Refinement} Mao et al.~\cite{mao2018catdet} present \textit{CaTDeT (Cascaded Tracking Detector)}, which applies a tracker as assistance for refining intermediate feature maps in video object detection. The illustration of the algorithm is shown in Fig.~\ref{fig:catdet}. The tracker provides information on the region of interest based on historical detection. It reduces the operation count by $5.1\times$ to $8.7\times$ with a slight delay.

\subsubsection{Selection in the Temporal Dimension}
Korbar et al. propose \textit{SCSampler}~\cite{korbar2019scsampler}, a lightweight model for clip sampling that can invoke recognition on only the most salient clips. The paper targets scenarios where the videos are untrimmed and long. They aim to reduce the high inference cost when every clip is executed on the clip classifier. They apply both visual and audio samplers and combine their saliency for prediction. The visual sampler proposes a learning-based clip-level saliency model that provides each clip a \textit{saliency score} between [0,1]. In particular, the model takes the input clip features that are fast to compute from the raw clip and have low dimensionality (lower than that of the classifier) to analyze each clip very efficiently. The clips with top-K \textit{saliency scores} are extracted with features by the classifier and aggregated together to form the representation for prediction. The sampler learning phase labels a \textit{saliency score} of video with $1$ if the action is contained in that clip; otherwise, it is labeled as $0$. The approach elevates the accuracy of an already state-of-the-art action classifier by $7\%$ and reduces its computational cost by more than 15 times. 

On the other hand, subset selection in the temporal dimension of the video can sometimes help improve task accuracy. This work typically involves the addition of some sophisticated mechanisms into the DNN. As a result, they often slow down the DNN rather than speed it up. But their goal is on the accuracy of the outputs rather than speed. A class of the work uses recurrent Neural Networks, such as Long Short Term Memory networks (LSTM), for capturing long-term dependencies in sequence data. They use them to generate compact video representations along with CNN activation maps~\cite{yeung2015endtoend, alwassel2017action, wu2018adaframe}. They mimic the behavior of humans viewing videos that often consist of a glimpse followed by several refinements. In addition, some work~\cite{wu2019multiagent} combines gated recurrent units (GRUs) and policy networks in reinforcement learning to adjust video sampling location accordingly at each time step.

\subsubsection{Selection in the Streaming Scenario}

\begin{figure}[ht]
  \includegraphics[width=0.5\textwidth,height=0.5\textheight,keepaspectratio]{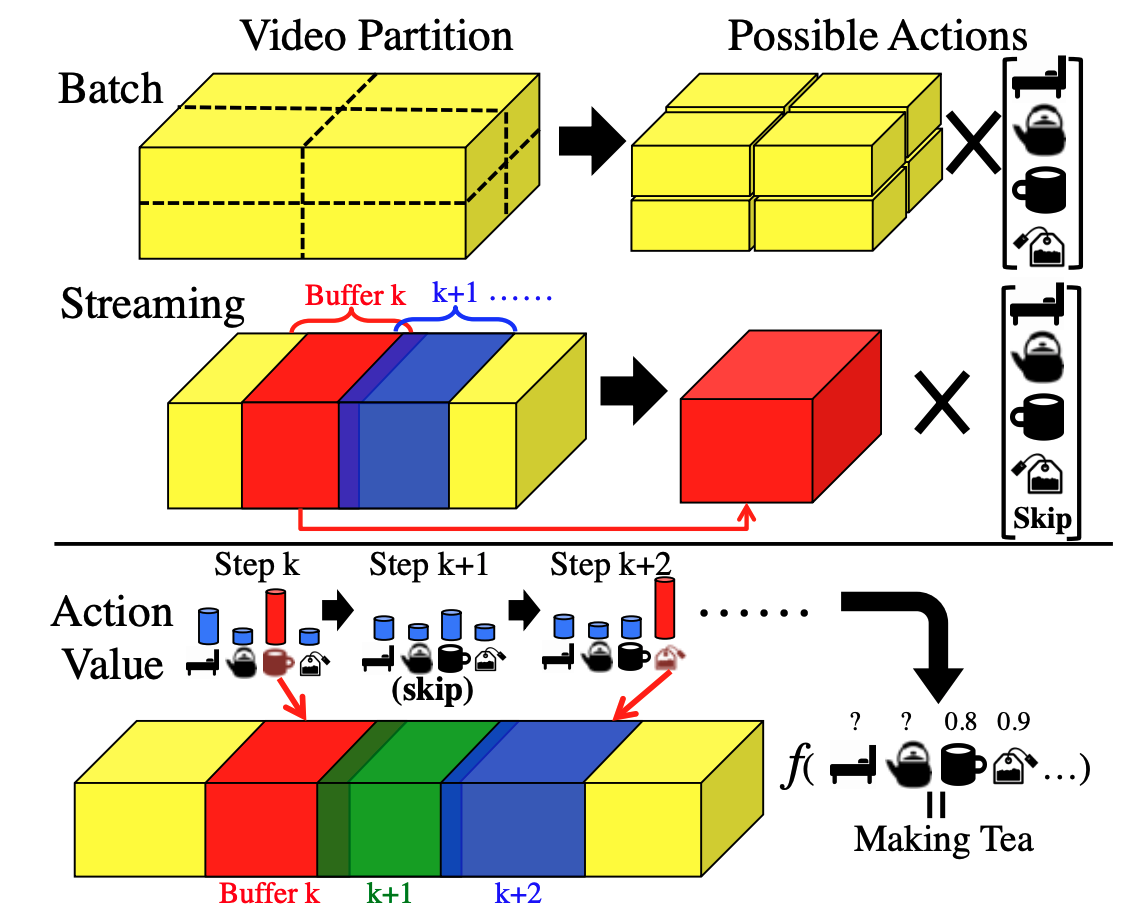}
  \caption{Action spaces. Top: In batch, the whole video is divided into sub-volumes, and actions are defined by the volume and object category to detect. Middle: In streaming, the video is divided into segments by the buffer at each step, and actions are the object category to detect in the buffer plus a “skip” action. Bottom: Our method learns a video-specific policy to select a sequence of valuable features to extract dynamically. Figure and description adapted from reference~\cite{su2016leaving}.}
  \label{fig:stone}
\end{figure}

The selection in a streaming scenario is more challenging. For each selection, the mechanisms need to make decisions on future frames based on historical records. Su et al.~\cite{su2016leaving} introduce an active mechanism that prioritizes \textit{"which feature to compute when"} to make timely action predictions. The core idea is to learn a policy that dynamically schedules the sequence of features to compute on selected frames of a given test video. The selection procedure can be invoked for either batch or streaming dimension. They formulate the problem with a Markov decision process (MDP) to learn the feature prioritization policy. The MDP sequentially selects frames with the most promising bag-of-objects or CNN features that can improve accuracy when combined with accumulated history results. Fig.~\ref{fig:stone} illustrates the action spaces explored in the mechanism. On two challenging datasets (Activities of Daily Living~\cite{pirsiavash2012detecting} and UCF-101~\cite{soomro2012ucf101}), their method provides significantly better accuracy than previous techniques under given computational budgets on two challenging datasets (Activities of Daily Living~ and UCF-101~).

\subsection{Temporal Propagation}
The neighboring frames in a video usually show some inherent continuity. In this type of optimization, the intermediate computation results of adjacent video frames such as feature maps, bounding boxes, or classification probability and confidence are propagated along the temporal dimension to reduce the need for re-computation on similar contents. Usually, some key frames are chosen, and the results are propagated from them to other frames. The optimizations are different in the representations used in propagation and update. The representations of frame features include \textit{Optical Flow}, \textit{Motion History Image (MHI)}, and \textit{Convolutional LSTMs}. To clarify, in \textit{Convolutional LSTMs}, frames are represented in normal CNN feature maps, but the corresponding information is updated via LSTM networks. 

\subsubsection{Using Optical flow}
\begin{figure}[ht]
  \includegraphics[width=0.65\textwidth,height=0.65\textheight,keepaspectratio]{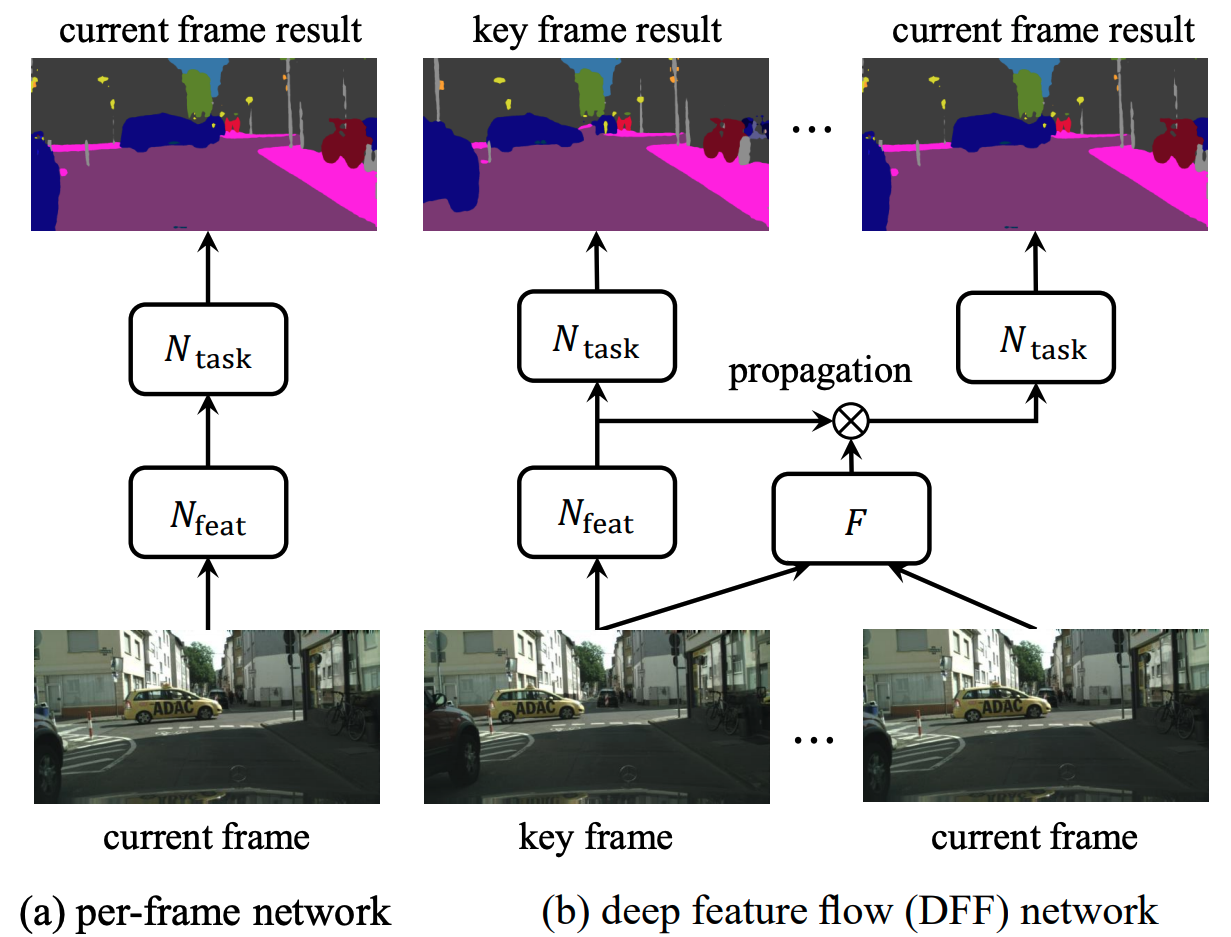}
  \caption{Illustration of temporal propagation in \textit{deep feature flow}. (a) Video recognition using per-frame network evaluation (b) The proposed \textit{deep feature flow}. Figures and descriptions adapted from reference~\cite{zhu2016deep}.}
  \label{fig:dff}
\end{figure}

Zhu et al.~\cite{zhu2016deep} propose a \textit{deep feature flow network (DFF)}. In \textit{DFF}, neighboring frames are propagated with feature maps via a flow field for partial updates. The updates are obtained by a flow estimation algorithm like \textit{SIFT Flow}~\cite{liu2008sift} or \textit{FlowNet}~\cite{dosovitskiy2015flownet}. Fig.~\ref{fig:dff} shows how information is propagated within neighboring frames. Compared to re-computing each frame through convolution layers, \textit{DFF} is $10\times$ faster with only a few percent accuracy loss. Subsequently, the team comes up with the \textit{flow-guided feature aggregation (FGFA)}~\cite{zhu2017flowguided} that associates and assembles the rich appearance information in consecutive frames to improve feature representation and accuracy. Afterward, the authors unify the solutions into a common framework~\cite{zhu2017high}, and achieve $77.8\%$ mAP score at a speed of $15.22$ frame per second for video object detection.

Kang et al.~\cite{Kang_2018} leverage motion-guided propagation for minor temporal information refinement across adjacent frames. The assistance of temporal information reduces the potential false negatives prediction when frames are processed per frame.

\subsubsection{Using Motion History Image}
\begin{figure}[ht]
  \includegraphics[width=\textwidth,height=\textheight,keepaspectratio]{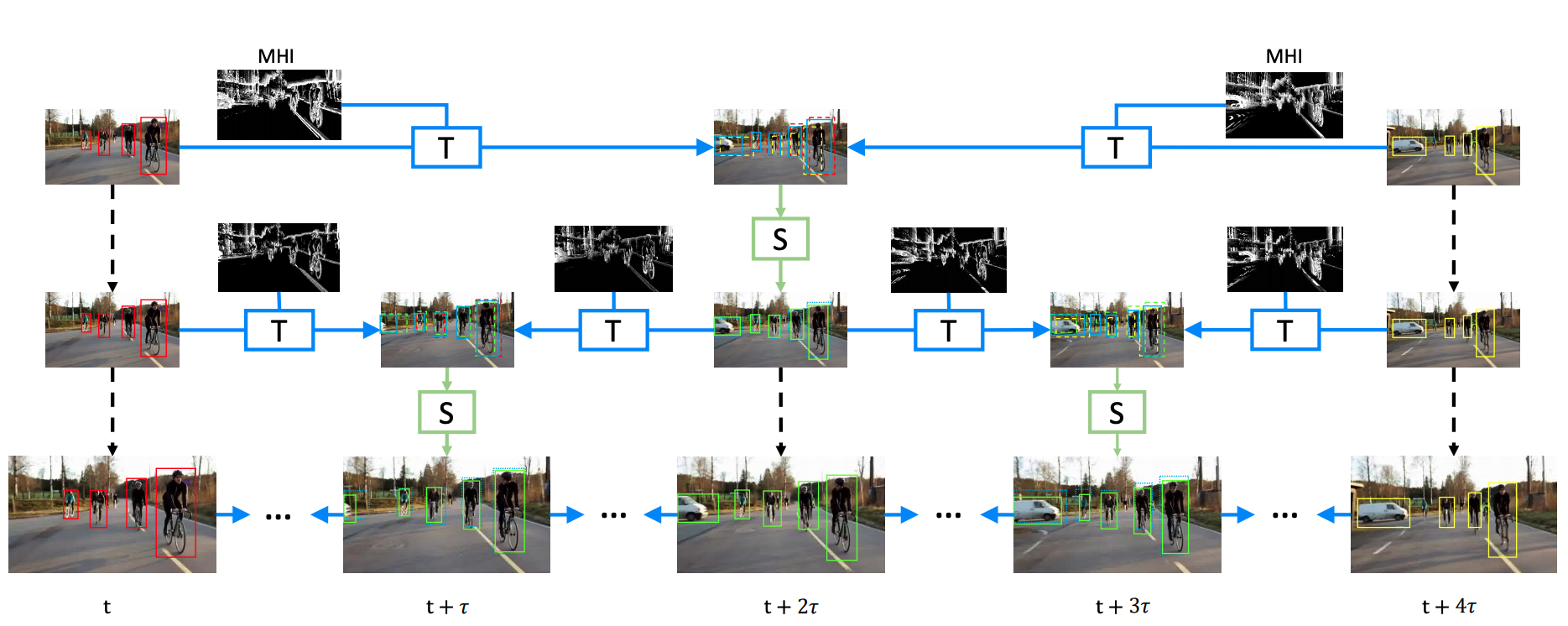}
  \caption{The \textit{Scale-Time Lattice}, where each node represents the detection results at a specific scale and time point, and each edge represents an operation from one node to another. In particular, the horizontal edges (in blue color) represent the temporal propagation from one time step to the next, while the vertical edges (in green color) represent the spatial refinement from low to high resolutions. Given a video, the image-based detection is only done at sparsely chosen key frames, and the results are propagated along a pre-defined path to the bottom row. The final results at the bottom cover all the time points. Figure and description adapted from reference~\cite{chen2018optimizing}.}
  \label{fig:stlattice}
\end{figure}

Unlike previous approaches that propagate key frames information at fixed intervals, Chen et al.~\cite{chen2018optimizing} propose \textit{Scale-Time Lattice} that selects the key frames adaptively. The \textit{Scale-Time Lattice} is a directed acyclic graph as shown in Fig.~\ref{fig:stlattice}. Inside the graph, the node stands for the bounding box detection result at a particular spatial resolution and time point; the edge represents an operation that performs temporal propagation or spatial refinement. During the execution, given a video input, the framework first applies expensive object detectors to the key frames selected sparsely and adaptively based on the object motion and scale. Next, within the lattice, those bounding boxes are propagated to intermediate frames and refined across scales (from coarse to fine) via cheaper networks. A \textit{Propagation Refinement Unit (PRU)} takes the detection results of two consecutive frames as input, propagates them to other frames, and re-scales them. The authors adopt \textit{Motion History Image (MHI)}~\cite{bobick2001recognition} since the motion representation computation is relatively cheap than optical flows. Overall, the proposed method yields 79.6 mAP at 20 FPS and 79.0 mAP at 62 FPS on ImageNet VID dataset. 

\subsubsection{Using Convolutional LSTMs}
\begin{figure}[ht]
  \includegraphics[width=0.5\textwidth,height=0.5\textheight,keepaspectratio]{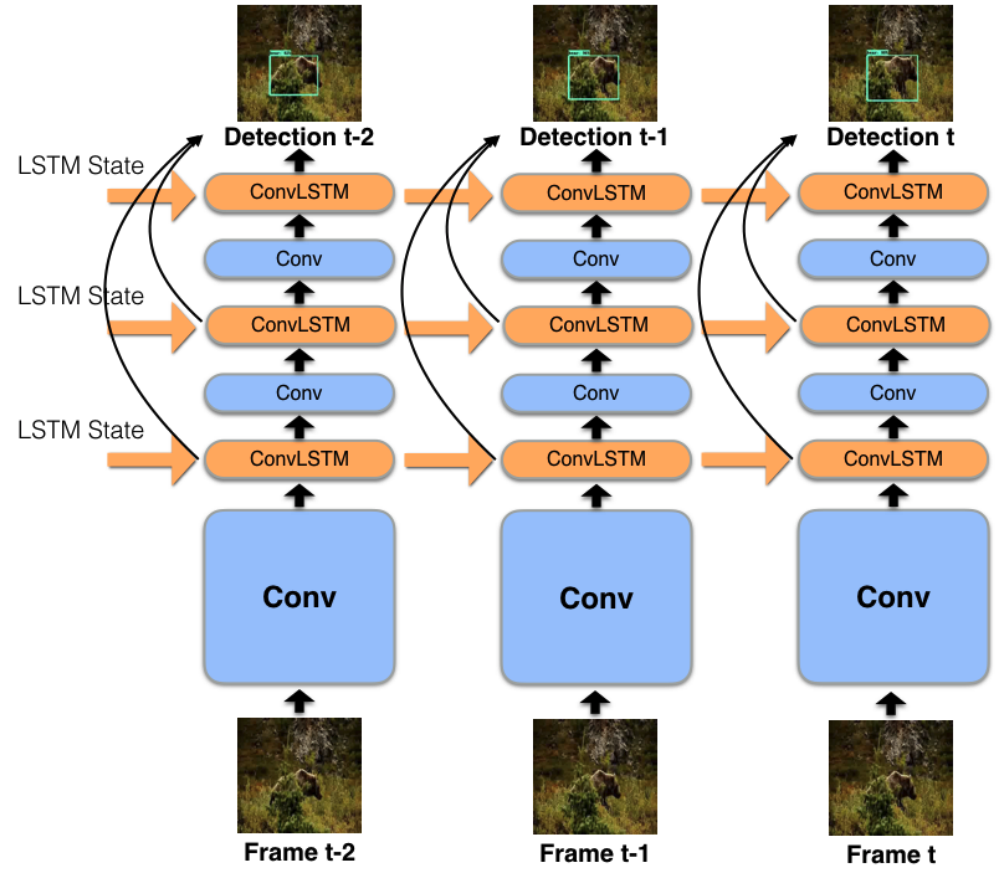}
  \caption{An example illustration of our joint LSTM-SSD model. Multiple Convolutional LSTM layers are inserted into the network. Each propagates and refines feature maps at a particular scale. Figure and description adapted from reference~\cite{liu2017mobile}.}
  \label{fig:lstmliu}
\end{figure}

Liu et al.~\cite{liu2017mobile} argue that temporal cues can be efficiently propagated through an LSTM network. The LSTM serves as an augmented structure to assist in refining temporal context for the generated features of each frame in the video. In the presented architecture in Fig.~\ref{fig:lstmliu}, a hypothesis feature map for each frame is generated by the CNN feature extractor and then fed into the LSTM to be fused with temporal context from previous frames. The output for that frame is a temporally-aware refined feature map. The experiment integrates the convolutional LSTMs with \textit{Single Shot Detector (SSD)}~\cite{liu2016ssd} and provides a quick inference speed with only 15 FPS on a mobile CPU.

\begin{figure}[ht]
  \includegraphics[width=0.75\textwidth,height=0.75\textheight,keepaspectratio]{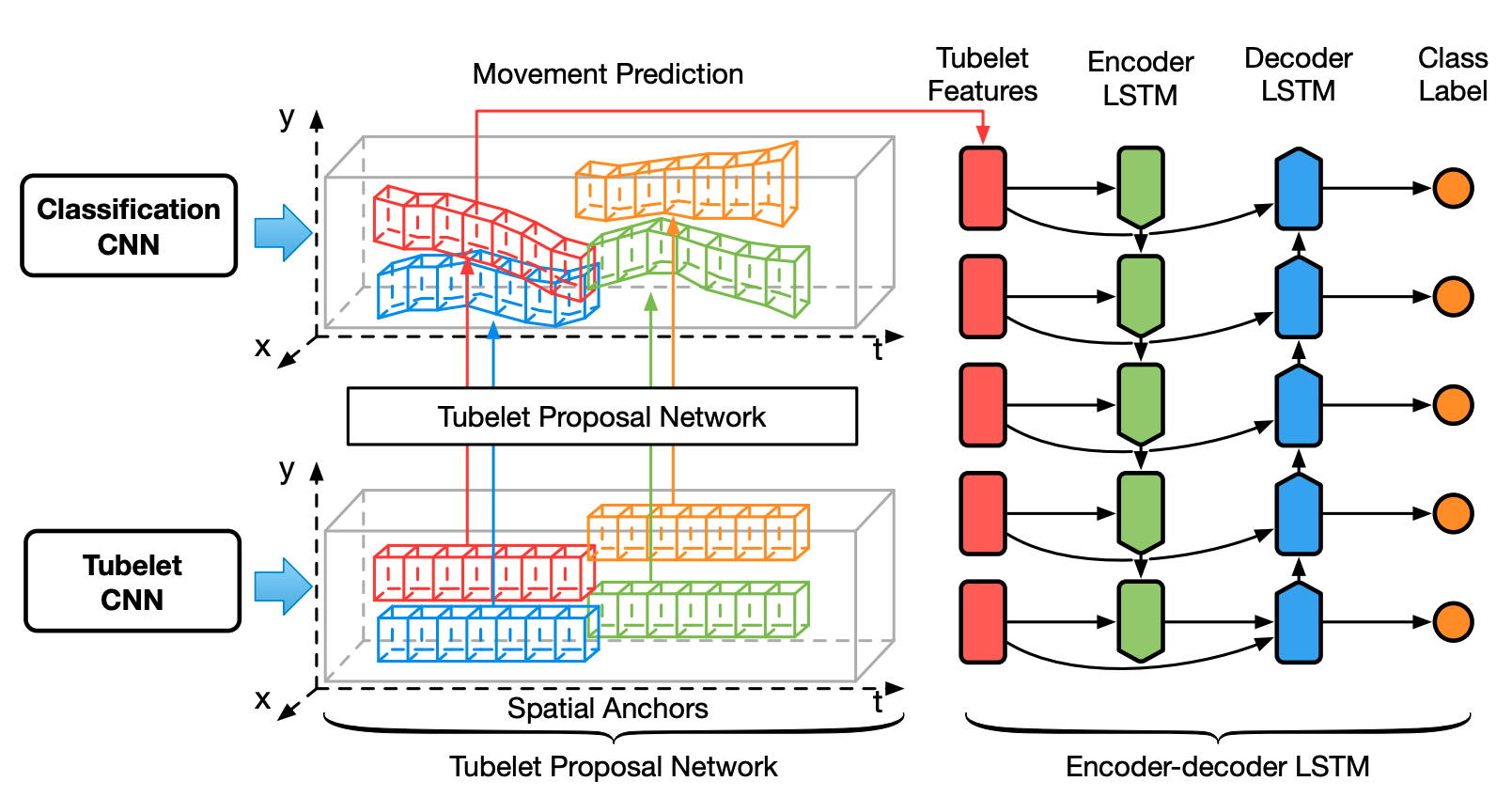}
  \caption{The proposed object detection system consists of two main parts. The first is a tubelet proposal network to generate tubelet proposals efficiently. The tubelet proposal network extracts multi-frame features within the spatial anchors, predicts the object motion patterns relative to the spatial anchors, and generates tubelet proposals. The gray box indicates the video clip, and different colors indicate the proposal process of other spatial anchors. The second part is an encoder-decoder CNN-LSTM network to extract tubelet features and classify each proposal box into different classes. The tubelet features are first fed into the encoder LSTM by a forward pass to capture the appearance features of the entire sequence. Then the states are copied to the decoder LSTM for a backward pass with the tubelet features. The encoder-decoder LSTM processes the whole clip before outputting class probabilities for each frame. Figure and description reference adapted from reference~\cite{Kang_2017}.}
  \label{fig:tcnn}
\end{figure}

Kang et al.~\cite{Kang_2017} leverage LSTMs instead of bounding box proposal refinement in video object recognition. The temporal information is propagated across each proposal tublet to improve accuracy. As shown in Fig.~\ref{fig:tcnn}, an encoder-decoder LSTM structure is placed after the \textit{Tublet Proposal Network}. It generates the output probability of each class label as a prediction result for each proposal.


\subsection{Data Redundancy based Optimizations on 3D CNNs} 
So far, we have discussed how the performance of 2D CNN is advanced with works that exploit redundancy, especially at the temporal dimension where contiguous frames involve much identical information. Most of the optimizations are for the performance of video object detection, which is rather heavy on computation as the DNN has an additional region proposal phase. Still, some DNNs enable the analytical capability for videos with either 3D CNN~\cite{6165309, tran2014learning} or two-stream models~\cite{NIPS2014_5353, Karpathy_2014_CVPR}. 
Unlike how 2D CNN is operated on video data, 3D CNN can end-to-end training and inference on videos. It is augmented with capturing semantics along the temporal dimension using stridden convolution filters.

Redundancy removal on video data for 3D CNN has been less explored at software-based optimizations. Some works~\cite{carreira2017quo, kpkl2019resource} propose to analyze the potential redundancy of the underlying network structures by progressively expanding the operator dimension from a 2D CNN basis. For example, Feichtenhofer et al.~\cite{Feichtenhofer_2020_CVPR} explore the 2D to 3D extension by considering the necessity of parameters or data in multiple dimensions. These include frame rate under fixed duration, temporal duration, spatial resolution, number of layers per residual stage, number of channels, and inner channel width in a residual block. These initiatives introduce another angle of viewing data redundancy on video DNNs. Future works offer prospective clues for advancing redundancy removal-based algorithms on deep models.

Still, existing works~\cite{wang2020edge, wang2019systolic, fan2017f, wang2017enhanced, shen2018towards, hegde2018morph, chen20193d} have developed performance optimizations for 3D CNN that extend the fruitful efforts on the 2D CNN counterparts in hardware-based optimizations. The promising outcomes lie in that computation along the temporal dimensions is only required on the deltas (value differences) between adjacent video frames. They effectively alleviate the memory consumption and inference time bottleneck on resource-constraint devices.





\section{Leverage Data Redundancy in Transformer Optimization for Text} \label{sec:text}
Text-based redundancy optimization can be applied to Recurrent Neural Networks (RNNs), the widely used deep learning model for texts before the invention of the Transformer-based model. Guan et al.~\cite{guan2021recurrent} propose to leverage context-free grammar (CFG) and a hierarchical compression algorithm to squeeze the repeated contexts in sequence prediction tasks. The optimization accelerates RNN inference up to a thousand times and expends the prediction scope without losing task accuracy. Similar optimization insights can be found in many recent Transformer models as well. Aside from reducing model execution time or increasing task-specific accuracy, many data redundancy-based optimizations in Transformer models are initially motivated by the need to scale up the capacity of the models. Each time, the model can only process a sequence limited in length (the number of token representations) for token-level text inputs, subject to the hardware capability. The problem is especially crucial when it is necessary to process the representations of a paragraph or an extended sequence at once. As a result, researchers have systematically identified redundancy within the raw data or intermediate data representations to scale the model correctly.

Unlike CNN optimizations which leverage data redundancy at input data or intermediate feature maps between convolution layers, optimizations on Transformers can occur at different stages of interim data representations. A sequence of text input to the transformer model goes through several encoding phases, each capturing an extra level of meaning. The stages focus on the initial sentences or sub-words input, the sub-words embeddings, the hidden states produced by positional encoding, multi-head attention, or the encoder layer as a whole (multi-head attention and feed-forward layer together). Data redundancy can exist in each of the phases. Existing studies leverage the redundancy of various insights. 

For the first stage, sub-word segmentation algorithms have been carried out with the motivation to remove redundancy in text data. \textit{Byte-Pair Encoding (BPE)}~\cite{sennrich2015neural} is one such case. For the second stage, redundancy may exist in input embeddings. Compression at the vector representation of word embeddings was introduced long before the wide use of Transformer models~\cite{adaptivecompression, acharya2018online, chen2016compressing}. For the remaining stages, there are optimizations that leverage redundancy at intermediate representation, \textit{"hidden state"} (a single sample output of a transformer layer; the unit is the same as activation map in CNN), of the model in general. Optimizations that leverage data redundancy in Transformer are classified based on how they exploit redundancy: clustering for reuse, skipping, and other ad-hoc sampling techniques. Table~\ref{tab:text} lists the surveyed work by their applications for better reference. 

\begin{table}[ht]
\caption{Summary of works on text data redundancy}
\label{tab:text}
\begin{tabular}{|p{2.5cm}|l|l|p{4.5cm}|p{4.5cm}|}
\hline
\textbf{Application} & \textbf{Paper} & \textbf{Code} & \textbf{Key Idea} & \textbf{Performance} \\ \hline
text summarization  & \cite{liu2018generating}        & \cite{liu2018generatinglink}        & \textit{T-DMCA} is a modified multi-head self-attention structure to reduce the memory footprint in self-attention layers                                                                     & possible to learn on sequence length of 12,000 on GPU (Nvidia P100)
                                                    \\ \hline
text classification & \cite{dai2020funneltransformer} & \cite{dai2020funneltransformerlink} & \textit{Funnel-Transformer} augments the transformer model with an encoder to gradually pooled the representation to reduce the sequence-length of the hidden states as the layer goes deeper & $1.3\times$-$1.5\times$ speedup on GPU/TPU        
                                                    \\ \cline{2-5} 
                    & \cite{goyal2020powerbert}       & \cite{goyal2020powerbertlink}       & \textit{PoWER-BERT} applies extract layers and learning-based retention configuration to retain only the key embedding vectors in the transformer model                                       & $4.5\times$ speedup on BERT-base models ($6.8\times$ with ALBERT) on GPU (Nvidia K80)
                                                    \\ \cline{2-5} 
                    & \cite{dalvi2020exploiting}      & \cite{dalvi2020exploitinglink}      & Exploit \textit{LayerSelector (LS)} and \textit{Correlation Clustering Feature Selection (CCFS)} to select the essence layers and features for transformer models                 & Evaluated on BERT and XLNET. Reduce forward pass parameters up to $47\%$. Reduce feature set to $4\%$ (sequence labeling), $<1\%$ (sequence classification) on GPU (Nvidia Titan X)
                                                    \\ \hline
question answering  & \cite{kitaev2019reformer}       & \cite{kitaev2020reformerlink}       & Propose \textit{Locality-sensitive Hashing Attention} and \textit{Reversible Transformer} to address data redundancy                                                              & Attention speed can maintain as low as 0.1-0.5 seconds per step when sequence length per batch scale from 32 to 32768s (baseline is 0.2-5 seconds per step) on GPU/TPU

                                                    \\ \hline
\end{tabular}
\end{table}

\subsection{Reuse via Clustering}
\begin{figure}[ht]
  \includegraphics[width=\textwidth,height=\textheight,keepaspectratio]{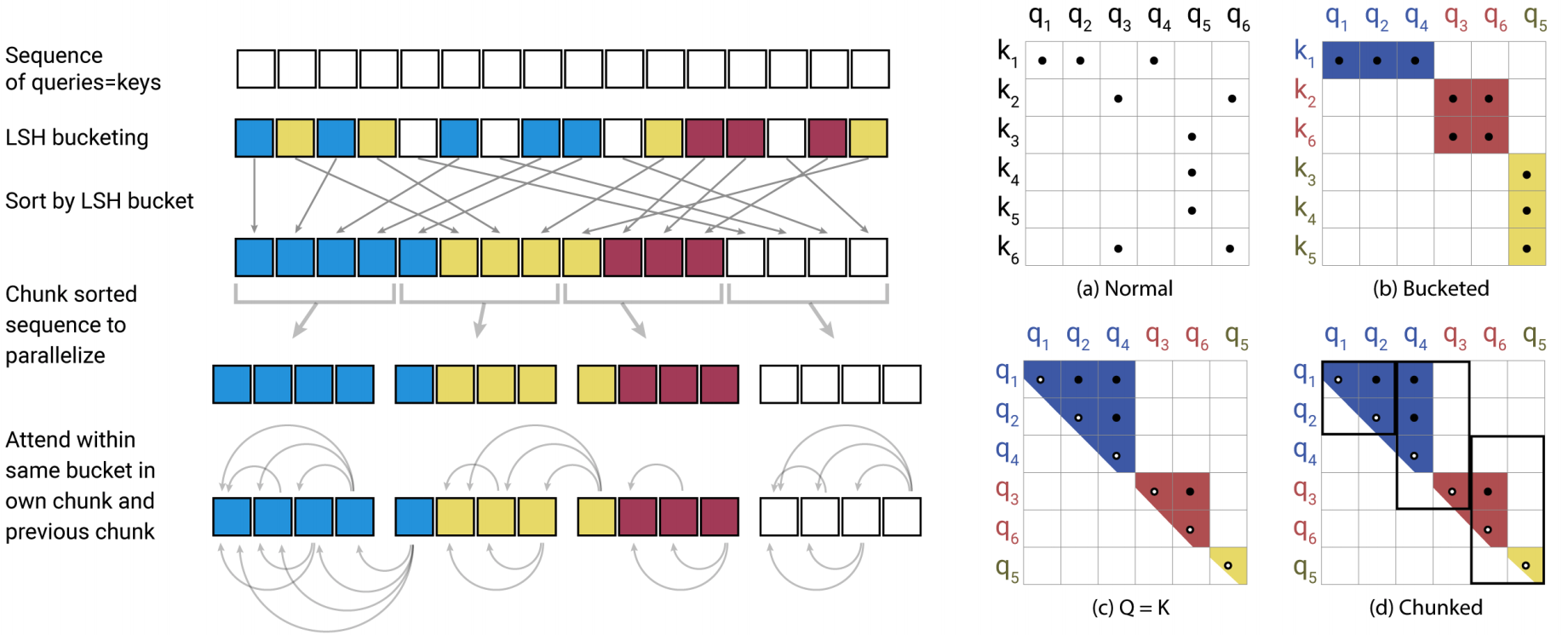}
  \caption{Locality-sensitive hashing attention in \textit{Reformer}. LSH Attention shows the hash-bucketing, sorting, and chunking steps and the resulting causal attentions. (a-d) Attention matrices for these varieties of attention. Figures adapted from reference~\cite{kitaev2019reformer}.}
  \label{fig:reformer}
\end{figure}

Kitaev and Kaiser et al. propose \textit{Reformer}~\cite{kitaev2019reformer} with two main contributions in improving the transformer model: \textit{Locality-sensitive Hashing Attention} and \textit{Reversible Transformer}. In particular, \textit{Locality-sensitive Hashing Attention} tackles data redundancy with a specific function design. The function uses an approximation in the \textit{scaled dot-product attention} in a transformer model. The \textit{scaled dot-product attention} is defined as $\frac{QK^{T}}{\sqrt{d_k}}V$, which $Q$, $K$, and $V$ are the query, key, and value matrices, respectively. The authors identify that the computation in the softmax function is dominated by key-query pairs that are in proximity. It indicates the calculation can pay attention to only a small subset of the closet key-query pairs. As exemplified in Fig.~\ref{fig:reformer}, \textit{locality-sensitive hashing (LSH)}~\cite{10.5555/645925.671516} is employed to first cluster the keys and queries into hash buckets. Queries and keys in each bucket are sorted to form chunks to avoid sparsity in the attention matrix. Afterward, the \textit{scaled dot-product attention} is substituted by allowing multiplication only between pairs in the same hash buckets (or in chunks in the attention matrix). Furthermore, the \textit{Locality-sensitive Hashing Attention} introduces a \textit{multi-round LSH attention} that enables multiple rounds of hashing to alleviate the problems that similar items may fall into different buckets after hashing. In the decoder of the transformer model, a masking mechanism is implemented to re-order the position indices by the same permutations that were previously applied to sort the key and query vectors.

\subsection{Skipping}
Output representations from attention layers in Transformer models have been recognized as containing much redundant information~\cite{michel2019sixteen, voita2019analyzing}. They have been addressed through ways of attention head pruning~\cite{michel2019sixteen, voita2019analyzing}. Recent studies have investigated the criteria for pruning neurons (activations) in Transformer models. Specifically, Gupta et al.~\cite{gupta2020compression2} suggest neurons are to be pruned when either the importance of that neuron is low or when the same activation multiplies with the same weights. The neuron importance function~\cite{sanh2019distilbert} is defined on entropy, output-weights norm, or the input-weights norm. 

\begin{figure}[ht]
  \includegraphics[width=0.65\textwidth,height=0.65\textheight,keepaspectratio]{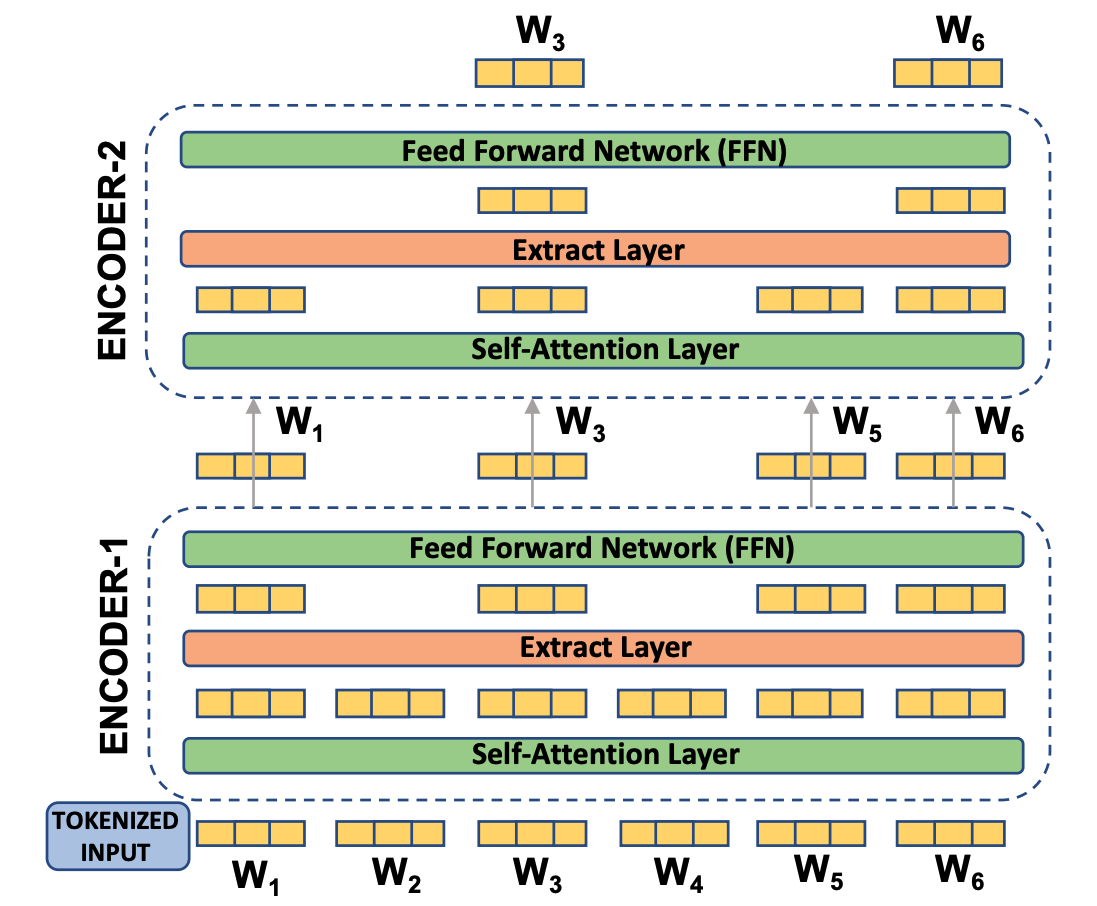}
  \caption{Illustration of \textit{PoWER-BERT}. Word-vector selection over the first two encoders. Here, $N = 6$, $l_{1} = 4$ and $l_{2} = 2$. The first encoder eliminates two word-vectors $w2$ and $w4$ with least significance scores; the second encoder further eliminates word-vectors $w1$ and $w5$. Figure and description adapted from reference~\cite{goyal2020powerbert}.}
  \label{fig:powerbert}
\end{figure}

The hidden state of a sequence of word vectors may contain repetitive information themselves. Goyal et al.~\cite{goyal2020powerbert} identify that the pair-wise cosine similarity between word vectors in hidden states increases as layers go progressively deeper (encode by more phases). They propose \textit{PoWER-BERT} to exploit the redundancy to reduce the inference time of the BERT model in text classification and regression tasks. As illustrated in Fig.~\ref{fig:powerbert}, \textit{PoWER-BERT} uses an \textit{Extract Layer} to manage the retention configuration after the \textit{Self-Attention Layer} and before the \textit{Feed Forward Network (FFN)} in each encoder layer of the Transformer model. The retention configuration can be set manually or set with parameters from training. The model is trained with a loss function proposed to obtain the learning-based design. This method avoids the exponential search space in retention configuration. Furthermore, they employ two kinds of strategies to retain the word representations: \textit{static} and the \textit{dynamic strategy}. The \textit{static strategy} keeps the word vectors at the same positions across all input sequences in the dataset. The \textit{dynamic strategy} retains the word vectors based on the attention scoring and a soft-extract layer. In the experiment, they obtain a $4.5\times$ inference time acceleration over the baseline BERT model with less than $1\%$ accuracy loss. The proposed technique is also compatible with the compression scheme such as \textit{ALBERT}~\cite{lan2019albert}. The combination of both compression methods achieves a $6.8\times$ inference time reduction with less than $1\%$ accuracy loss.

Dalvi et al.~\cite{dalvi2020exploiting} propose an efficient transfer learning method that exploits \textit{LayerSelector (LS)} and \textit{Correlation Clustering Feature Selection (CCFS)} to select the essence layers and features required during transformer model execution. There are three main steps in the selection:  (1) \textit{LayerSelector (LS)} uses the layer-classifier to select the lowest layer that maintains oracle performance. (2) \textit{Correlation Clustering} filters out redundant neurons given the hidden state's output of the previous layer. (3) \textit{Feature Selection (FS)} selects a minimal set of neurons required to reach optimum performance on the given task.

For the first step, the authors analyze task-specific layer-level redundancy by training linear probing classifiers~\cite{shi2016does, belinkov2017neural} on each layer $l_{i}$ as \textit{layer-classifier}. They use the \textit{layer-classifier} to select the lowest layer that maintains oracle performance (maintaining $99\%$ performance). In \textit{LS}, a concentration of all layers until the chosen layer is used. 

For the following two steps, the \textit{CCFS} exploits data redundancy by a combination of clustering and subset selection:
\begin{itemize}
\item \textit{Correlation Clustering (CC)}~\cite{bansal2004correlation}: For every length-N sequence vector representation, the product-moment correlation between each of the two features is calculated. This gives a $N$-by-$N$ matrix $corr(x,y)$ representing the correlation between $fx$ and $fy$. The correlation value ranges from $-1$ to $1$, which provides a relative scale to compare any two neurons. The distance metric $cdist(x, y)$ between these two features is defined by $1-|corr(x,y)|$. A hyper-parameter $ct$ defines the maximum distance between any two features to be considered as a cluster.
\item \textit{Feature Selection (FS)}: It identifies a minimal set of neurons that match the oracle performance. The \textit{Linguistic Correlation Analysis}~\cite{dalvi2019one} ranks the neurons concerning a downstream task.
\end{itemize}

Overall, the work finds out that up to $85\%$ and $95\%$ neurons are redundant in BERT and XLNet~\cite{yang2019xlnet} respectively. In the experiment, the proposed method accelerates sequence labeling tasks by $2.8\times$ and $6.2\times$ on BERT and XLNet, respectively. While for sequence classification tasks, the average speedups are $1.1\times$ and $2.8\times$ correspondingly.

\subsection{Other Sampling}
\begin{figure}[ht]
  \includegraphics[width=0.75\textwidth,height=0.75\textheight,keepaspectratio]{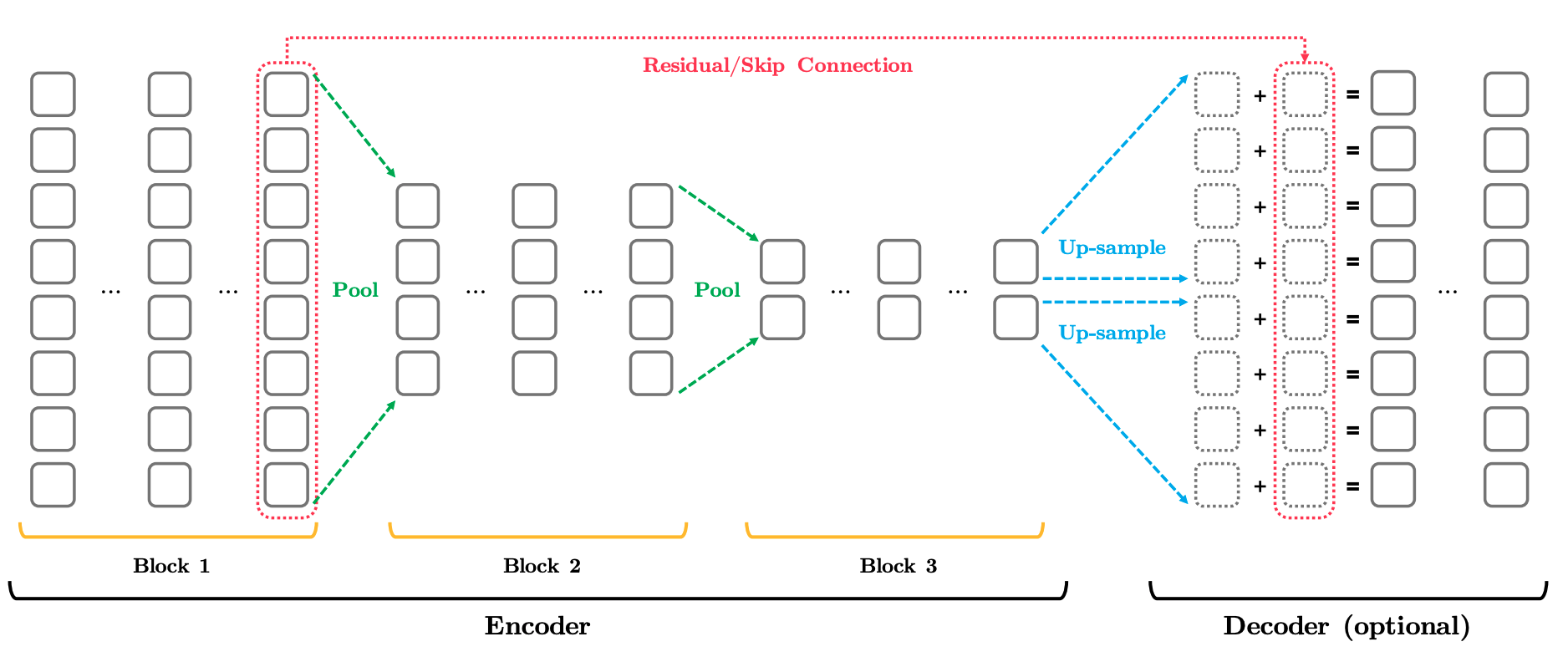}
  \caption{Illustration of Pooling in Funnel-Transformer. The encoder on the left consists of several blocks of consecutive Transformer layers. Within each block, the sequence length of the hidden states always remains the same. But when going from a lower-level block to a higher-level block, the size of the hidden sequence is reduced by performing a particular type of pooling along the sequence dimension. The right part showcases the up-sampling to re-construct the arrangement of the original length. Figure and description adapted from reference~\cite{dai2020funneltransformer}.}
  \label{fig:funneltransformer}
\end{figure}

Humans implicitly use prior linguistic knowledge to merge nearby tokens or words in paragraphs. This, in turn, forms more extensive semantic phrases or units to understand sentences or paragraphs in a general form. Inspired by that, Dai et al.~\cite{dai2020funneltransformer} have proposed \textit{Funnel-Transformer}. As shown in Fig.~\ref{fig:funneltransformer}, the \textit{Funnel-Transformer} is equipped with an additional encoder to gradually pool the representation to reduce the sequence length of the hidden states as the layer goes deeper. The sub-modules have an extra residual connection and layer normalization operation in the \textit{Self-Attention Layer} and the \textit{Feed Forward Network (FFN)} of the Transformer model. The decoder is applied to reconstruct the full-sequence of representation when the token-level outputs are required, such as pretraining.

\begin{figure}[ht]
  \includegraphics[width=0.75\textwidth,height=0.75\textheight,keepaspectratio]{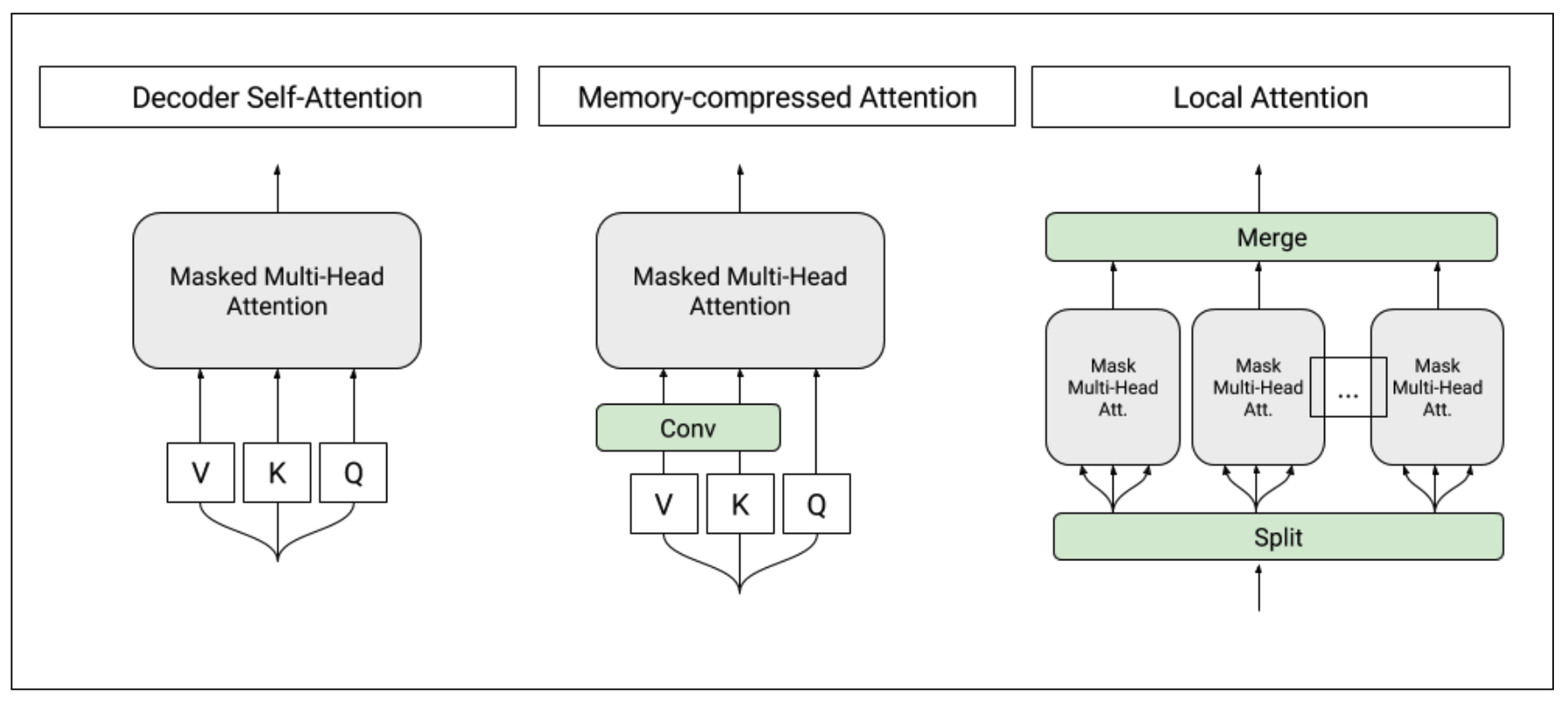}
  \caption{Transformer Decoder with Memory-Compressed Attention(T-DMCA). Every attention layer takes a sequence of tokens as input and produces a sequence of similar length as the output. Left: Original self-attention as used in the transformer decoder. Middle: Memory-compressed attention, which reduces the number of keys/values. Right: Local attention, which splits the sequence into individual smaller sub-sequences. The sub-sequences are then merged to get the final output sequence. Figure and reference adapted from reference~\cite{liu2018generating}.}
  \label{fig:generating}
\end{figure}

To achieve the generation of Wikipedia-styled summarization over multi-documents, Liu et al.~\cite{liu2018generating} introduce a compressive scheme for long text sequences. The \textit{Transformer Decoder with Memory-Compressed Attention(T-DMCA)} is a modified multi-head self-attention structure to reduce the memory footprint via limiting the dot product between the query and key of a \textit{Self-Attention Layer}. The mechanism consists of the \textit{Memory-Compressed Attention} and the \textit{Local Attention} as shown in Fig.~\ref{fig:generating}. The \textit{Memory-Compressed Attention} reduces the number of keys and values by using a stridden convolution (for sampling). The \textit{Local Attention} divides a sequence of tokens into blocks of similar length of tokens (they use 256 tokens) to allow constant attention memory cost per block regardless of sequence length. After feeding the sub-sequence to their corresponding \textit{Multi-head Attention} to capture local information, the method merges the results to get the final output sequence. For both \textit{Local} and \textit{Memory-Compressed Attention}, masking is cast to prevent the queries from attending to future keys and values. Finally, the design enables the model to process 3$\times$-longer sequences than the default model.

\section{Future Research Directions} \label{sec:future}
The many previous studies have clearly shown the significant benefits of exploitation of data redundancy in Deep Learning. In creating the summary of the many explorations from various aspects, we have recognized the broad coverage of the existing innovations, but at the same, observed some open problems and several research directions worth pursuing in the future.

(i) \textbf{Principled understanding of the relations between data redundancy exploitation and accuracy.} Data redundancy exploitation could cause changes in the output of a DNN because the removed data are often not guaranteed to be redundant. In most cases, it faces the risks of degrading the accuracy, but in some cases, as mentioned earlier~\cite{chin2019adascale, korbar2019scsampler, su2016leaving}, it may also improve the accuracy for its noise removal effects. There is a lack of principled understanding of how data redundancy elimination affects model accuracy. The accuracy of a DNN model can be affected by existing variance related to data. Dataset itself may already contain bias over DNN models~\cite{torralba2011unbiased}. Data preprocessing, such as data augmentation~\cite{shorten2019survey, cubuk2019autoaugment}, makes the discussion even more complicated. Previous work has either resorted to empirical tuning to reach a satisfying point or used some heuristic methods to rate the salience of some data. An open question is how to achieve a principled understanding of the impact and transform the knowledge into principled data redundancy exploitation techniques. Some rigorous studies may help.

(ii) \textbf{Enhancing interpretability.} Related to the first direction, the second direction that may be worth exploring is interpretability, which refers to both the interpretability of neuron representation and the interpretability of DNN in general. Utilizing evidence in data redundancy has led to some initial success in improving the interpretability of CNN’s feature representation, contributing to the CNN model pruning processes. For example, Li et al.~\cite{li2019exploiting} propose a \textit{kernel sparsity and entropy indicator (KSE)} to quantify the importance of each pixel in the activation map to provide feature-agonistic guidance in weight compression. Identifications of data redundancy also benefit the interpretation of the importance of each word in text processing. For example, Dalvi et al.~\cite{dalvi2019one} present \textit {Linguistic Correlation Analysis} and \textit{Cross-model Correlation Analysis} to recognize the relative importance of a neuron in a Transformer model. Their method offers an ablation view on the saliency of words in the paragraphs to guide future representation development. These several studies show some promise in exploring data redundancy for the interpretability of deep learning. As interpretability is important for connecting Deep Learning with domain understanding, more efforts are worthwhile to develop further. 

(iii) \textbf{Other data types and models.} A large portion of the prior work on data redundancy is about image and video data. There is still some room left for more explorations on these data types. An example is learning from point cloud~\cite{guo2020deep} where data redundancy has not yet been explored. But in comparison to images and videos, redundancy in text data is much less explored, as Transformer is more recently developed than the more mature models used in images and videos. More explorations are especially in demand to better understand data redundancy in texts. Moreover, besides the three types of data, there are other types of data on which DNN is also playing an important role. These data include graphs, scientific simulation results, genes, computer programs, and so on. Correspondingly, the special properties of these data have prompted the development of some new DNN models, such as Graph Convolutional Networks (GCN)~\cite{wu2020comprehensive}. The differences in data properties, DNN models, and learning tasks suggest that some research could be fruitful in understanding and exploiting data redundancy in these areas.

(iv) \textbf{Relations among optimizations and systematic solutions.} One of the questions that have not received much attention is the relations among the many kinds of optimizations on data redundancy elimination, as well as their relations with model optimizations. On images, for instance, as Section~\ref{sec:image} and Table~\ref{tab:overview} show, there are many ways to exploit data redundancy from various angles. Can these optimizations be used together? Is it worth doing that? Would there be some conflicts besides that they may all affect the model accuracy? How to use them together in the best way? More ambitiously, is it possible to build up frameworks that automatically apply one or more kinds of data redundancy exploitation techniques on a given dataset and learning task? How about the interplay with model optimizations? Answers to these questions may significantly advance the current understanding of data redundancy exploitation and tap into the potential better.

(v) \textbf{Synergy with DNN accelerator designs.} Yet another direction worth looking into is the synergy between the many data redundancy exploitation techniques and DNN hardware designs. In modern application-specific integrated circuit (ASIC) design, dataflow analysis~\cite{10.1145/3352460.3358252, ben2019stateful, MLSYS2021_c9e1074f} has been recognized as a crucial consideration. Some work on edge devices has tried to address memory capacity issues by exploiting data redundancy (e.g., through data compression)~\cite{mocerino2019energy, salamat2018rnsnet, hegde2018ucnn, wang2020edge, wang2019systolic, fan2017f, wang2017enhanced, shen2018towards, hegde2018morph}. But in general, there is no systematic understanding of the relations of the many data redundancy exploitation techniques and the designs of DNN hardware accelerators. On the one hand, do those techniques stay profitable when DNN runs on hardware accelerators? On the other hand, 
can hardware accelerator designs gain some efficiency benefits by adopting some of the ideas in those techniques? Answers to these questions could lead to the novel synergy between the two strands of efforts.

\section{Conclusion} \label{sec:conclusion}
Machine learning is based on two pillars, models and data. As an essential part of deep learning, data plays an important role. Making the best use of data determines the quality, speed, efficiency, and interpretability of machine learning. This survey summarizes the efforts in the research community in detecting and leveraging data redundancy in a variety of dimensions, introduces the first known taxonomy to categorize the existing techniques, and points out a set of directions yet to explore. As the landscape of machine learning evolves continuously at an unprecedented speed, we hope that this survey can provide a one-stop resource for researchers to attain a quick understanding of state of the art and open issues, and hence benefit the industry practitioners and research community in selecting existing solutions for solving a problem at hands or creating new solutions to advance this critical field further.

\bibliographystyle{ACM-Reference-Format}
\bibliography{sample-base}

\end{document}